\documentclass[times, final, 10pt]{elsarticle}
\usepackage{algorithm}
\usepackage{algorithmic}
\usepackage{comment}
\usepackage{amsmath}
\usepackage{amssymb}
\usepackage{multirow}
\usepackage{booktabs}
\usepackage{array}
\usepackage{subcaption}
\usepackage[table]{xcolor}
\usepackage{newfloat}
\usepackage{listings}
\usepackage{adjustbox}
\makeatletter

\newcommand{\Rmnum}[1]{\expandafter\@slowromancap\romannumeral #1@}
\makeatother

\biboptions{numbers,sort&compress}
\usepackage{amssymb}

\usepackage{url}
\usepackage{amsmath,amsfonts}
\usepackage{algorithmic}
\usepackage{algorithm}
\usepackage{array}
\usepackage{textcomp}
\usepackage{stfloats}
\usepackage{verbatim}
\usepackage{graphicx}

\usepackage{multirow}
\usepackage{booktabs}
\usepackage{color}
\usepackage{tabularx}
\usepackage{amsmath} 
\usepackage{enumitem}
\usepackage{caption}

\usepackage[hidelinks]{hyperref}
\usepackage{orcidlink}

\usepackage{pifont}
\usepackage{booktabs}
\usepackage{multirow}
\usepackage{threeparttable}
\usepackage{xcolor}
\usepackage{graphicx}
\usepackage{booktabs}
\usepackage{multirow}
\usepackage[table]{xcolor}

\usepackage{lineno}
\journal{Pattern Recognition}

\begin{document}

\begin{frontmatter}

\title{Exploiting Stability-Plasticity Asymmetry in Pretrained Detectors for Incremental Object Detection}

\author[1]{Songze Li}
\author[1]{Qixing Xu}
\author[1,2,3]{Tonghua Su\corref{cor1}}
\author[4,5]{Xu-Yao Zhang}
\author[1]{Zhongjie Wang}
\author[1]{Yunzhe Li}
\cortext[cor1]{Corresponding Author. Email: thsu@hit.edu.cn}
\affiliation[1]{
    organization={Harbin Institute of Technology},
    city={Harbin},
    country={China} 
}
\affiliation[2]{
    organization={Guangdong Laboratory of Artificial Intelligence and Digital Economy (SZ)},
    city={Shenzhen},
    country={China}
}
\affiliation[3]{
    organization={Chongqing Research Institute of HIT},
    city={Chongqing},
    country={China}
}
\affiliation[4]{
    organization={State Key Laboratory of Multimodal Artificial Intelligence Systems, Institute of Automation of Chinese Academy of Sciences},
    city={Beijing},
    country={China}
}
\affiliation[5]{
    organization={School of Artificial Intelligence, University of Chinese Academy of Sciences},
    city={Beijing},
    country={China}
}

\begin{abstract}

Pretrained model-based incremental object detection (PTMIOD) leverages the rich detection priors of pretrained detectors to learn new categories incrementally while preserving detection ability on previously learned ones. 
Existing methods mainly exploit pretrained detectors as a whole, without explicitly distinguishing which components should remain stable and which require plastic adaptation. 
In this paper, we revisit PTMIOD from a component-wise stability-plasticity perspective. 
Our analysis of pretrained DETR-based detectors reveals a clear asymmetry: localization heads preserve transferable geometric priors across tasks, whereas classification-related representations require greater plasticity to handle new categories, especially in cross-domain scenarios where downstream data deviate from the pretraining domain. 
Based on this finding, we propose a selective adaptation and retention  framework that freezes explicit localization heads to preserve localization stability, while adapting transformer representations with parameter-efficient fine-tuning and updating classification heads for classification-oriented plasticity. 
To alleviate classification-side forgetting, we pioneer the use of pseudo-feature replay in PTMIOD and design Quality-aware Gaussian Feature Replay, which estimates reliable class-wise feature distributions from high-quality matched object features and replays sampled pseudo features to maintain old-class decision boundaries. 
Since continual adaptation can shift the feature space and undermine replayed distributions, we further develop Two-stage Consistent Distillation to align teacher and student representations at both proposal generation and refinement stages.
Extensive experiments on COCO, VOC, and TT100K show that our method achieves state-of-the-art performance under both in-domain and cross-domain incremental detection settings, demonstrating a favorable balance between old-class retention and new-class adaptation.

\end{abstract}   


\end{frontmatter}


\section{Introduction}
\label{sec:intro}
Recent advances in deep learning have significantly advanced object detection systems 
\cite{ren2015faster,carion2020end}.
Most existing detectors are designed under a static learning paradigm, assuming all target categories are pre-defined during training. 
However, real-world applications require continuous adaptation to new categories over time. 
Simply fine-tuning models on new data inevitably leads to catastrophic forgetting \cite{goodfellow2013empirical,mccloskey1989catastrophic}—a critical issue where models rapidly lose previously learned knowledge. 
Conversely, retraining models with combined old and new data is often infeasible due to privacy constraints and prohibitive computational costs.
To this end, incremental object detection (IOD)~\cite{DONG2023109488} has been proposed and extensively studied, aiming to address the challenges of continuously learning new object while maintaining the detection performance on previously learned ones.
 
Pretrained model-based incremental object detection (PTMIOD) has recently emerged as a promising direction for IOD. 
Before the rise of large-scale pretrained detectors, most IOD methods typically trained detectors from scratch in the incremental setting.
To mitigate catastrophic forgetting, they rely on techniques such as knowledge distillation~\cite{peng2020faster}, pseudo-labeling~\cite{liu2023continual}, and exemplar replay~\cite{yuyang2023augmented}.
With the rapid development of pretrained models, recent studies have begun to explore PTMIOD, aiming to leverage the strong generalization ability of pretrained detectors for continual learning. 
Among them, two-stage DETR-based detectors such as DINO~\cite{zhang2023dino} have become a popular choice due to their strong representation and localization capabilities. 
Building upon these pretrained detectors, recent PTMIOD methods~\cite{bhatt2024preventing,an2025parameterized} introduce prompt-based adaptation and achieve encouraging results.

Despite the progress of PTMIOD, existing methods still lack an in-depth understanding of how stability and plasticity are distributed across different components of pretrained detectors. 
Most existing methods treat the pretrained detector as a whole and do not explicitly identify which components should be preserved for stability and which components should be adapted for plasticity during incremental learning. 
This limitation becomes particularly critical in \textit{cross-domain} scenarios, where the downstream data differ substantially from the pretraining domain. 
Although existing PTMIOD methods have shown encouraging performance on \textit{in-domain} settings, such as incremental learning on datasets whose categories and visual concepts are aligned with the pretraining data, their adaptability to cross-domain scenarios remains insufficiently explored. 
As illustrated in Fig.~\ref{fig:main_figure}, representative PTMIOD methods can perform well on in-domain objects, but struggle to recognize cross-domain objects such as traffic signs, indicating that the generalization ability inherited from pretrained detectors does not uniformly transfer across localization and classification components. 
This observation raises a key question for PTMIOD:  whether different components of a pretrained detector require the same stability-plasticity treatment during incremental learning? 
In addition, many existing IOD and PTMIOD methods rely on pseudo-labeling to preserve old-class knowledge~\cite{gupta2022ow,kim2024vlm,liu2023continual}. 
Such methods implicitly depend on the presence of old-class objects in future-task images, which restricts their applicability in practical incremental scenarios.

Motivated by these observations, we revisit PTMIOD from a component-wise stability-plasticity perspective. 
Our analysis shows a component-wise stability-plasticity asymmetry: localization heads of pretrained DETR-based detectors exhibit strong transferable stability across domains, while classification-related representations are more sensitive to new categories. 
Based on this finding, we propose a selective adaptation and retention framework.  Specifically, we preserve localization-oriented priors by freezing the explicit localization heads, while applying parameter-efficient fine-tuning (such as LoRA) to transformer representations and updating classification heads to improve classification plasticity. 
For old-class retention, we pioneer the use of pseudo-feature replay in PTMIOD and design Quality-aware Gaussian Feature Replay (QGFR), which estimates old-class feature distributions from high-quality matched object features and replays sampled pseudo features to preserve old-class decision boundaries, thereby avoiding the restrictive assumption of pseudo-labeling that old-class objects must appear in future-task images. 
Since continual adaptation can shift the feature space and weaken the reliability of stored Gaussian distributions, we further develop Two-stage Consistent Distillation (TCD) to stabilize feature-space replay by enforcing teacher-student consistency in two-stage DETR detectors. 
Extensive experiments on COCO~\cite{lin2014microsoft}, VOC~\cite{everingham2010pascal}, and TT100K~\cite{zhu2016traffic} demonstrate that our method achieves state-of-the-art performance under both in-domain and cross-domain incremental detection settings.

The main contributions of this paper are summarized as follows:

\begin{figure}[t]
    \centering 
    \includegraphics[width=\linewidth]{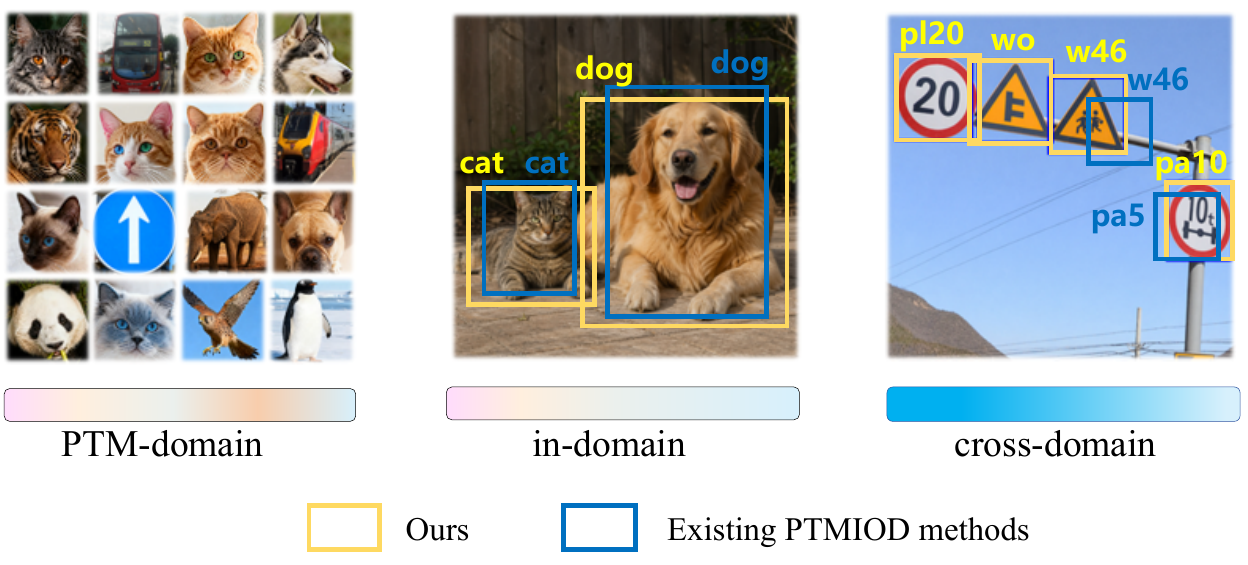}
\caption{Illustration of the PTM-domain and a comparison of detection results on in-domain vs. cross-domain tasks.
\textbf{Left:} Examples illustrating the diversity of the PTM-domain (e.g., Objects365 \cite{shao2019objects365}).
\textbf{Middle \& Right:} Comparison of detection results. While both our method and representative PTMIOD methods perform well on the in-domain task (middle), existing PTMIOD methods often fail on the cross-domain task (right), where our method remains effective.
}
    \label{fig:main_figure} 
\end{figure}

\begin{itemize}

    \item We reveal a component-wise stability-plasticity asymmetry in pretrained DETR-based detectors and propose a selective adaptation and retention framework that preserves localization stability while enhancing classification plasticity with LoRA-based adaptation.

    \item  To our knowledge, we are the first to introduce pseudo-feature replay into PTMIOD with explicit feature-space stabilization. Specifically, we propose Quality-aware Gaussian Feature Replay to preserve old-class decision boundaries without stored exemplars, and develop Two-stage Consistent Distillation to mitigate feature-space drift under continual adaptation. 

    \item Extensive experiments on COCO, VOC, and TT100K demonstrate that our method achieves state-of-the-art performance under both in-domain and cross-domain IOD settings.

\end{itemize}

\section{Related Works}
\subsection{DETR-based Detectors} 
DETR~\cite{carion2020end} introduces end-to-end object detection with transformers and inspired a family of DETR-based detectors. 
Deformable DETR~\cite{zhu2020deformable} improves DETR with deformable attention and introduces a two-stage variant that uses encoder proposals as decoder queries.
DAB-DETR~\cite{liu2022dabdetr} improves query formulation, while DN-DETR~\cite{li2022dn} enhances training stability through denoising training.
Building on these advances, DINO~\cite{zhang2023dino} develops a stronger two-stage DETR design with contrastive denoising and mixed query selection, achieving substantially better detection performance.
Recent works such as Grounding DINO~\cite{liu2024grounding} and T-Rex2~\cite{jiang2024t} further extend DETR-based detectors to the pretrained setting, suggesting that pretrained DETR-based detectors are becoming general and transferable detection foundations.

\subsection{Incremental Learning}
Incremental learning aims to enable models to acquire new knowledge while retaining previously learned information. 
Traditional methods are commonly grouped into three categories: rehearsal-based~\cite{castro2018end,rebuffi2017icarl}, parameter-isolation~\cite{mallya2018piggyback,mallya2018packnet}, and regularization-based approaches~\cite{kirkpatrick2017overcoming,li2024continual}. 
Rehearsal-based methods mitigate forgetting by replaying past data or stored exemplars, while parameter-isolation methods allocate separate parameters or modules to different tasks.
Regularization-based methods preserve prior knowledge by constraining model updates, for example through parameter importance estimation or knowledge distillation.

Recently, pretrained models have become an important foundation for incremental learning due to their strong generalization ability~\cite{zhou2024continual}. Among PTM-based approaches, prompt-based adaptation~\cite{YUAN2026113704,FENG2026112276} is a representative parameter-efficient solution. It adapts pretrained models to new tasks by learning a small set of task-specific prompts while keeping the main model unchanged. Another line of work uses pseudo-feature replay. For example, SLCA~\cite{zhang2023slca} and DUKAE~\cite{LI2026113860} mitigate catastrophic forgetting by replaying generated features. Specifically, these methods model the feature statistics of previous classes and then generate pseudo features for replay. The generated features are used together with new-task data to reduce forgetting without storing original samples. Furthermore, CoGaMiD~\cite{zhucontinual} extends this strategy to class-incremental semantic segmentation by modeling old-class feature distributions with Gaussian mixtures. 
Most pseudo-feature replay methods are designed for image classification or semantic segmentation, while extending them to object detection remains underexplored and challenging, as detection requires features that support both classification discrimination and localization awareness.

\subsection{Incremental Object Detection}
Incremental object detection (IOD) extends continual learning to object detection, requiring models to learn novel categories over time while preserving both classification and localization performance on previously learned objects.
A distinctive property of IOD is the old--new co-occurrence phenomenon: current-task images may contain unlabeled objects from previous tasks, which can provide useful signals for preserving old knowledge.
Existing methods often exploit this property through knowledge distillation~\cite{peng2020faster,cermelli2022modeling} or pseudo-labeling~\cite{gupta2022ow,liu2023continual,kim2024vlm}.
Knowledge distillation constrains the current model to match the outputs or intermediate features of the previous model, while pseudo-labeling treats old-class predictions generated by the previous model as additional supervision.
More recently, PTMIOD has emerged as a promising direction, aiming to exploit the strong generalization ability of pretrained detectors for continual learning.
MD-DETR~\cite{bhatt2024preventing} introduces task-specific prompts for each incremental task and selects prompts using features extracted by the pretrained detector.
To address the matching and task confusion issues in the prompt pool of MD-DETR, P$^2$IOD~\cite{an2025parameterized} directly generates prompts based on features extracted by the pretrained detector.
Both of these methods rely on pseudo-labeling, and their prompt mechanisms usually depend on the quality of features extracted by the original pretrained detector.

\section{Preliminaries}
\label{sec:third}
\textbf{Incremental Object Detection.}
Object detection aims to both locate and classify objects within an image.
Given an input image \(I\), a detector \(\mathcal{M}_{\theta}\) produces a set of predictions \(\mathcal{M}_{\theta}(I)=\{(\hat{\mathbf{b}}_i,\hat{c}_i)\}_{i=1}^N\), where each bounding box \(\hat{\mathbf{b}}_i \in \mathbb{R}^4\) and each class label \(\hat{c}_i \in C\) comes from a predefined set of classes.
Typically, the detector is trained on full dataset \({D}\) assuming all object classes \(C\) are available simultaneously.
Our work is situated in the setting of class-incremental object detection, which extends this conventional detection framework to an incremental learning setting. 
For a sequence of tasks \(1,...t,...,T\) with corresponding datasets \({D_1},...,{D_t}...,D_T\), let the class set for task \(t\) be defined as \(C_t\) such that \(C_t \cap C_s= \emptyset\), for \(t \neq s\).
The full dataset and class set can be expressed as \(D=\bigcup_{t=1}^T D_t\), \(C=\bigcup_{t=1}^T C_t\).
Each dataset \(D_t\) consists of images that may contain objects from both current and previous categories, but only objects in the current category set \(C_t\) are annotated; all other objects are treated as background during training.
The goal of IOD is to update model incrementally from \(\mathcal{M}_{t-1}\) to \(\mathcal{M}_{t}\) by learning new classes in \(C_{t}\) using \(D_{t}\) without access to previous datasets \(\{D_1,...,D_{t-1}\}\), while maintaining detection performance on \(\bigcup_{s=1}^{t-1}C_s\).

\textbf{Revisiting DINO.}
Following~\cite{an2025parameterized}, we use DINO pretrained on Objects365~\cite{shao2019objects365} as the baseline to study the stability-plasticity trade-off in PTMIOD.
DINO is a DETR-based detector that extends standard DETR with a two-stage detection paradigm; the first and second stages refer to the encoder-side and decoder-side predictions, respectively.
Given an input image \(I \in \mathbb{R}^{H_0 \times W_0 \times 3}\), DINO first performs an encoder-based proposal generation stage.
Specifically, the backbone network \(\mathcal{B}\) extracts multi-scale image features, which are further encoded by the transformer encoder \(\mathcal{E}\) as
\begin{equation}
\mathbf{Z}^1 = \mathcal{E}(\mathcal{B}(I; \theta_b); \theta_e) \in \mathbb{R}^{HW \times d},
\end{equation}
where \(d\) denotes the feature dimension.
Each encoded spatial feature \(\mathbf{z}_i^1 \in \mathbf{Z}^1\) is associated with a normalized reference point 
\(\mathbf{p}_i=(p_{ix},p_{iy})\in[0,1]^2\), which is determined by its position on the feature map.
Given the reference point, the first-stage localization module \(\mathcal{G}^1\) and classification head \(\mathcal{F}^1\) generate a proposal-score pair at each spatial location:
\begin{equation}
(\hat{\mathbf{b}}_i^1, \hat{\mathbf{s}}_i^1)
=
\left(
\mathcal{G}^1(\mathbf{z}_i^1, \mathbf{p}_i; \theta_{g_1}),
\mathcal{F}^1(\mathbf{z}_i^1; \theta_{f_1})
\right),
\end{equation}
where \(\hat{\mathbf{b}}_i^1\) and \(\hat{\mathbf{s}}_i^1\) denote the proposal box and classification scores at location \(i\), respectively.
The proposals are ranked by their maximum classification scores over all categories, and the top-\(k\) indices are selected as
\begin{equation}
\mathcal{I}=\mathrm{Top}\text{-}k\big(\{\max_c s_{i,c}^1\}_{i=1}^{HW}\big),\quad\mathcal{I} \subseteq \{1,\dots,HW\}.
\end{equation}
The corresponding proposal boxes are used as positional queries for the decoder:
\begin{equation}
Q^{\text{pos}}=\left\{\hat{\mathbf{b}}_{i}^1 \mid i \in \mathcal{I}\right\}.
\end{equation}
Together with the learnable content queries \(Q^{\text{con}}\in \mathbb{R}^{k \times d}\), the transformer decoder \(\mathcal{D}\) attends to the encoder representations and produces proposal-aware object features:
\begin{equation}
\mathbf{Z}^2 = \mathcal{D}(\mathbf{Z}^1, Q^{\text{pos}}, Q^{\text{con}}; \theta_d) \in \mathbb{R}^{k \times d}.
\end{equation}
The decoder features are then fed into the second-stage classification head \(\mathcal{F}^2\) and localization module \(\mathcal{G}^2\) to produce the final detection results:
\begin{equation}
(\hat{\mathbf{b}}_j^2, \hat{\mathbf{s}}_j^2)
=
\left(
\mathcal{G}^2( \mathbf{z}_j^2,\mathbf{q}_j^{\text{pos}}; \theta_{g_2})),
\mathcal{F}^2(\mathbf{z}_j^2; \theta_{f_2})
\right),
\quad j=1,\ldots,k,
\end{equation}
where \(\mathbf{q}_j^{\text{pos}}\) denotes the \(j\)-th positional query in \(Q^{\text{pos}}\).

During training, DINO supervises both the selected first-stage proposals and the final decoder predictions using a DETR-style set prediction objective.
Specifically, the predictions are matched to ground-truth objects through Hungarian matching~\cite{kuhn1955hungarian} and optimized with classification and box-regression losses, including focal loss~\cite{lin2017focal}, GIoU loss~\cite{rezatofighi2019generalized}, and L1 loss.
DINO further introduces a denoising objective to improve query learning from perturbed ground-truth objects.
The overall training objective is summarized as
\begin{equation}
\mathcal{L}_{\text{DINO}} =
\mathcal{L}_{\text{s1}} +
\mathcal{L}_{\text{s2}} +
\mathcal{L}_{\text{dn}},
\end{equation}
where \(\mathcal{L}_{\text{s1}}\), \(\mathcal{L}_{\text{s2}}\), and \(\mathcal{L}_{\text{dn}}\) denote the first-stage proposal loss, the second-stage decoder loss, and the denoising loss, respectively.
\section{Method}
\begin{figure}[t]
    \centering
    \includegraphics[width=\textwidth]{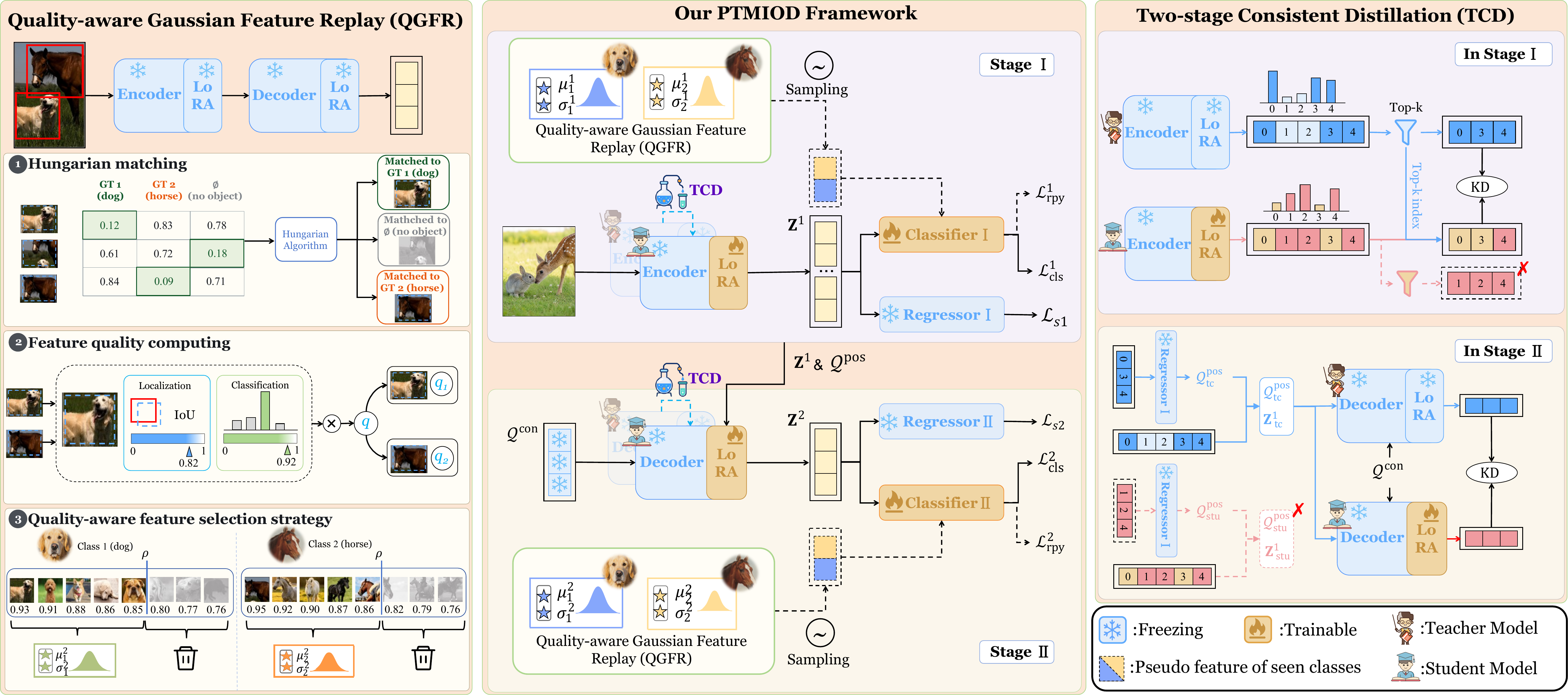}
    \caption{Overview of the PTMIOD framework.
    \textbf{Left:} Feature collection and Gaussian modeling for Quality-aware Gaussian Feature Replay (QGFR), illustrated with decoder features; the encoder follows the same procedure.
    Matched features are collected through Hungarian matching, filtered by their quality scores, and then used to fit class-wise Gaussian distributions.
    \textbf{Middle:} The framework performs continual adaptation with LoRA and supports old-class retention through QGFR, while TCD stabilizes the feature space.
    \textbf{Right:} Two-stage Consistent Distillation (TCD) alleviates the structural mismatch in standard distillation by enforcing consistency between the teacher and the student at both stages of DINO.}
    
    \label{fig:framework}
    
\end{figure}

In this section, we first analyze the stability and plasticity of different components in pretrained DINO detectors.
The analysis reveals a clear asymmetry between localization and classification: the localization heads exhibit strong transferable stability,  while classification-related representations are more sensitive to new categories and downstream domain shifts.
Based on this finding, we propose a selective adaptation and retention framework that preserves stable localization while enabling flexible adaptation of classification features.
To retain previously learned categories during this adaptation process, we introduce QGFR and further develop TCD to stabilize the feature space on which QGFR depends.
QGFR models previously learned categories with class-wise feature distributions through quality-aware Gaussian estimation and replays pseudo-features sampled from these distributions during subsequent tasks.
TCD stabilizes feature representations in two-stage DETR-based detectors, thereby maintaining the reliability of replayed old-class distributions during continual adaptation.
The overall framework is illustrated in Fig.~\ref{fig:framework}.

\subsection{Stability-Plasticity Asymmetry of Pretrained Detector}
\label{sec:analysis}

PTMIOD requires a pretrained detector to learn new categories without destroying the detection knowledge inherited from pretraining. 
Uniformly adapting the detector can disrupt transferable priors, whereas excessive preservation can limit the plasticity needed for new-category learning, especially under cross-domain scenarios. 
Since object detection involves both localization and classification, a key question is whether these two components require the same stability-plasticity treatment during incremental learning.
We argue that they do not. 
Localization mainly relies on geometric and spatial priors, while classification depends more on semantic discriminability in the feature space. 
We therefore analyze pretrained DETR-based detectors by separating localization and classification under both in-domain and cross-domain settings.

\textbf{Generalizable Localization Priors.}
We first examine whether the localization ability of the PTM generalizes across different downstream datasets.
Our hypothesis is that the localization path mainly captures geometric priors, such as object shape and spatial structure, rather than heavily relying on domain-specific semantics.
If true, freezing the modules that jointly determine box localization while training only the classification head should preserve localization quality, even when the downstream domain differs from the pretraining domain. 
Specifically, we freeze all modules initialized from the pretrained detector that are responsible for localization, including the backbone, transformer, box regression heads, and the first-stage classification head \(\mathcal{F}^1\), while updating only the second-stage classification head \(\mathcal{F}^2\).
We then evaluate class-agnostic recall at an IoU threshold of 0.5. 
In this metric, all predicted and ground-truth boxes are treated as a single class, so the result reflects localization performance independently of category prediction.
For comparison, we also report a fully fine-tuned baseline under the same evaluation protocol.
As shown in Table~\ref{tab:recall}, the frozen setting incurs only small recall drops on VOC, COCO, and TT100K, indicating that the localization ability of the PTM generalizes well across both in-domain and cross-domain datasets.\footnote{Additional results on more datasets are provided in Appendix A.}

\begin{table}[t]
\centering
\caption{Class-agnostic Recall@50 results of the model on the VOC, COCO, and TT100K datasets under two training settings. Recall@50 is evaluated at an IoU threshold of 0.5, with all predicted and ground-truth boxes treated as a single class. Lower recall drop indicates better generalization of localization ability.}
\label{tab:recall}
\setlength{\tabcolsep}{5pt}
\renewcommand{\arraystretch}{1.0}
\footnotesize
\begin{tabular}{lccc}
\toprule
\textbf{Dataset} & \textbf{Upper Bound} & \textbf{Frozen Loc.} & \textbf{Drop \(\downarrow\)}  \\
\midrule
VOC    & 99.5 & 99.3 & 0.2 \\
COCO   & 99.3 & 99.0 & 0.3 \\
TT100K & 99.7 & 97.3 & 2.4 \\
\bottomrule
\end{tabular}
\end{table}

\begin{figure}[t]
    \centering

    \captionsetup[subfigure]{font=scriptsize, skip=1pt}
    \captionsetup{font=small, skip=3pt}

    \newcommand{\tsneimg}[1]{%
        \includegraphics[
            width=0.95\linewidth,
            trim={2pt 2pt 2pt 2pt},
            clip
        ]{#1}%
    }

    \begin{subfigure}[t]{0.43\textwidth}
        \centering
        \tsneimg{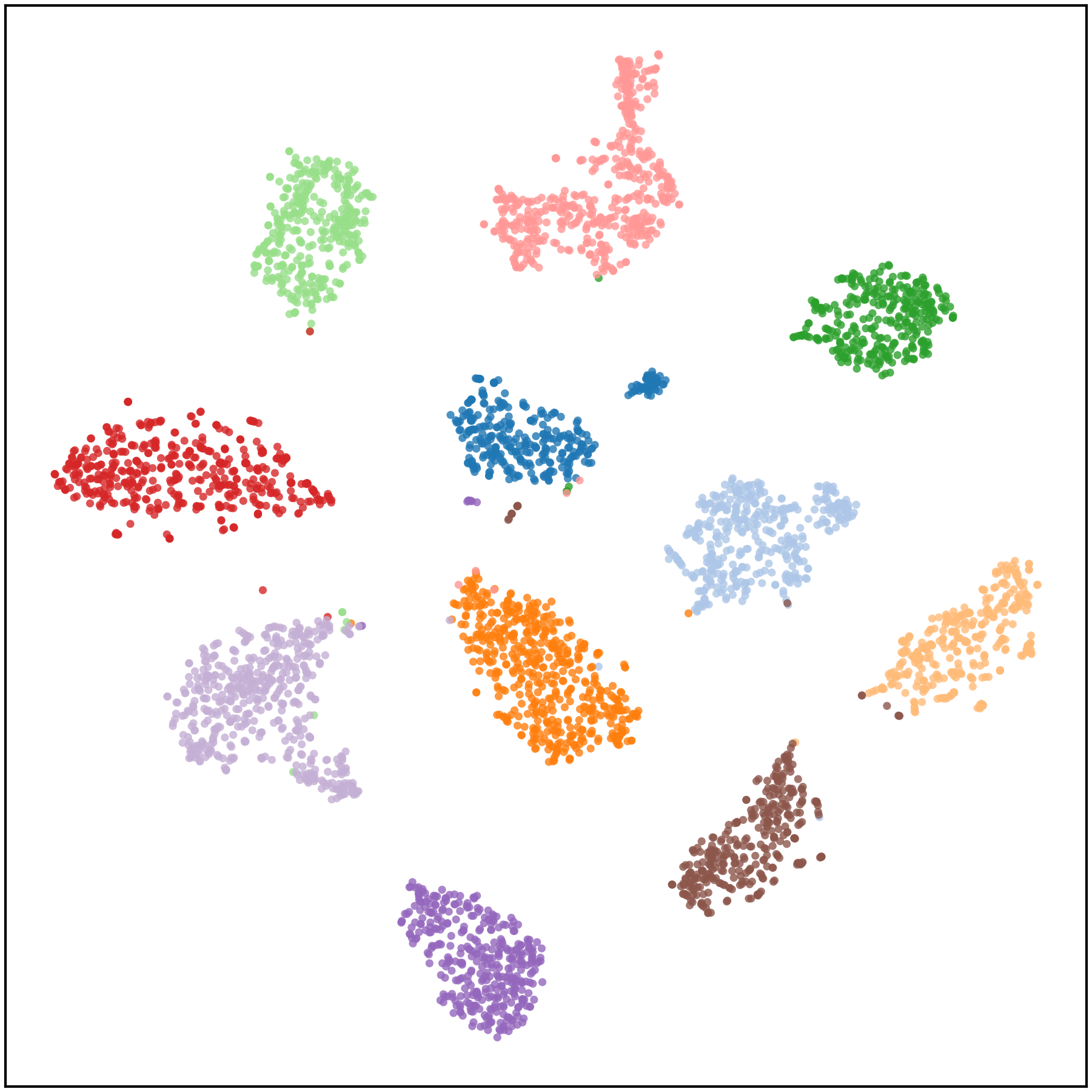}
        \caption{VOC Original}
        \label{fig:tsne_voc_cls}
    \end{subfigure}
    \hspace{0.03\textwidth}
    \begin{subfigure}[t]{0.43\textwidth}
        \centering
        \tsneimg{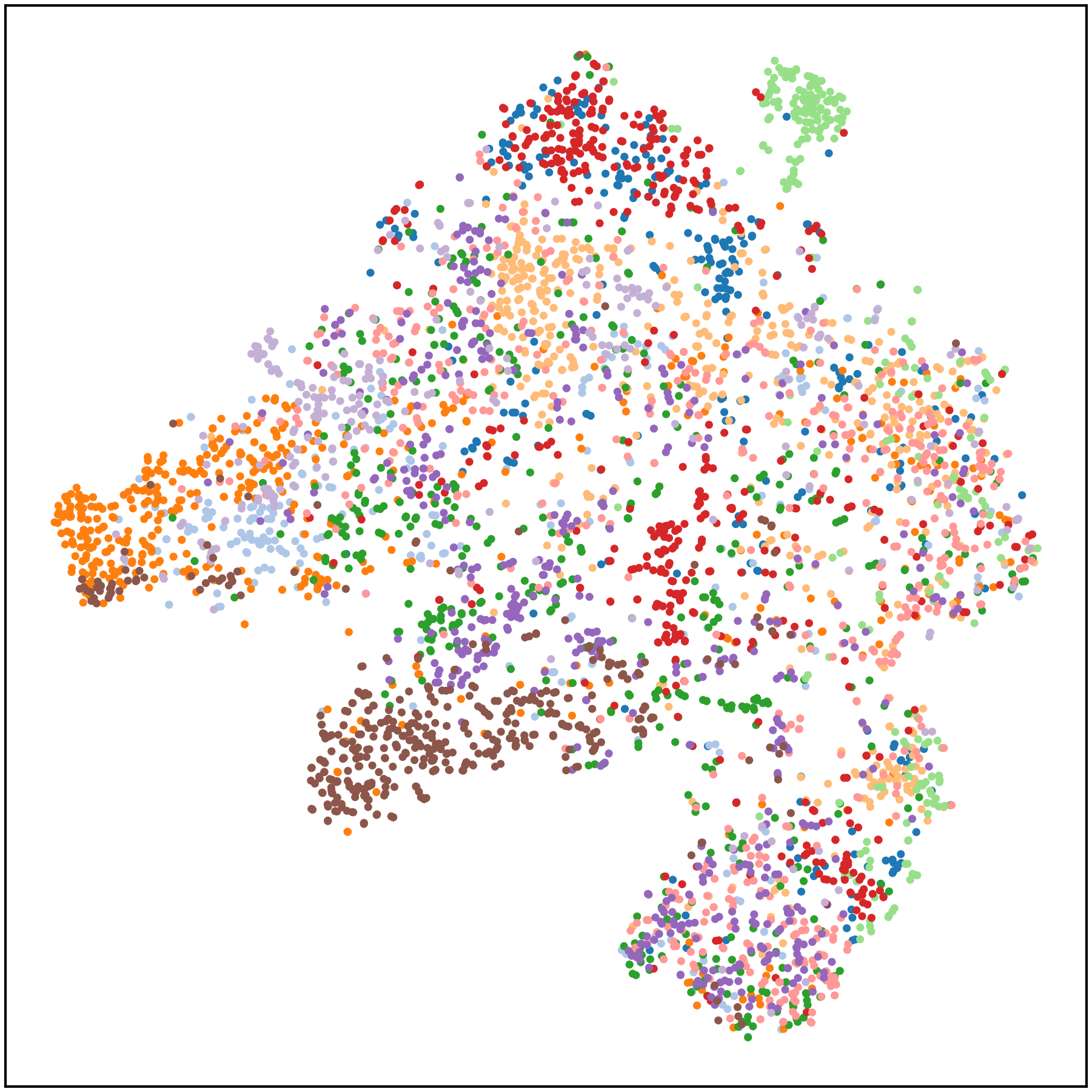}
        \caption{TT100K Original}
        \label{fig:tsne_tt100k_cls}
    \end{subfigure}

    \vspace{0.35em}

    \begin{subfigure}[t]{0.43\textwidth}
        \centering
        \tsneimg{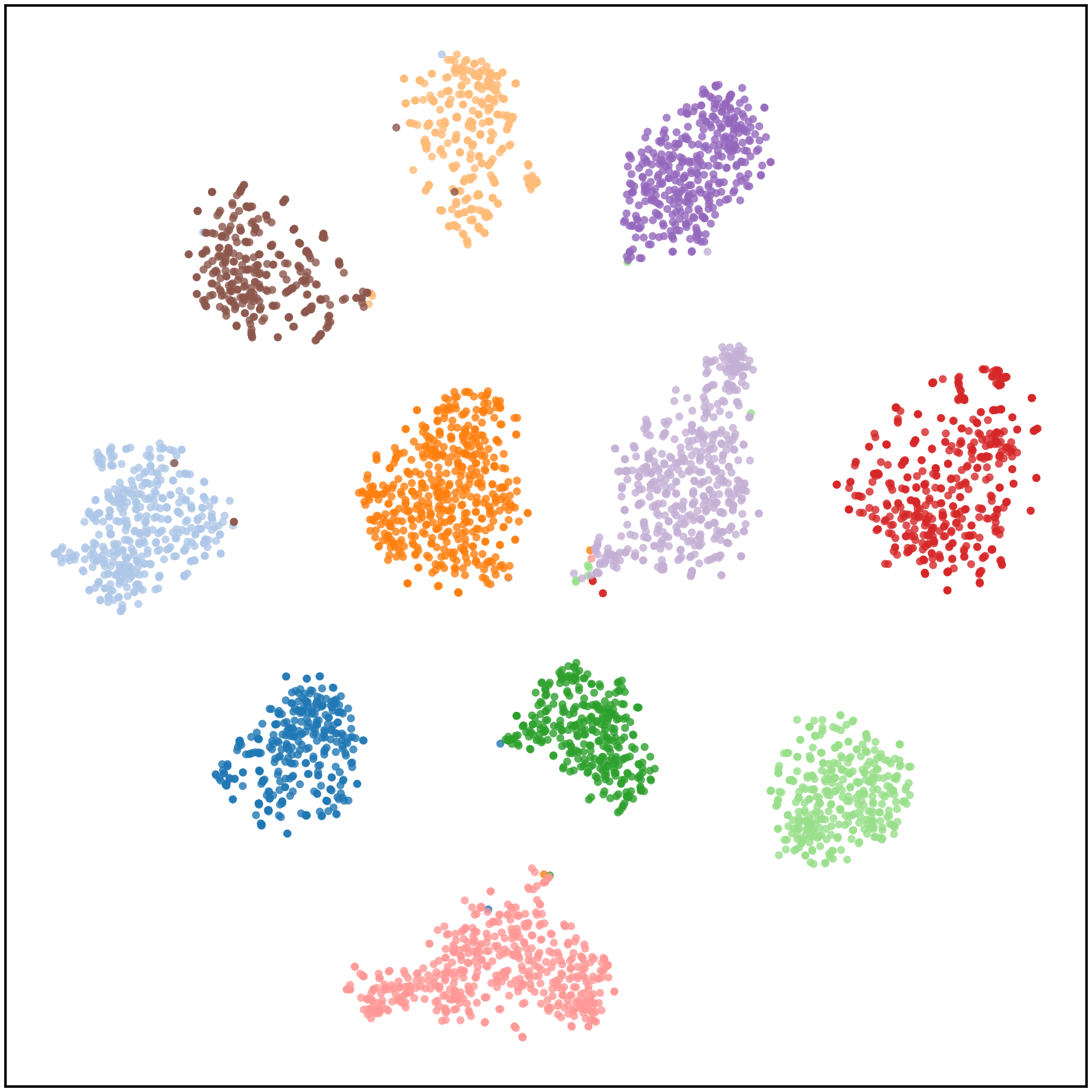}
        \caption{VOC w/ LoRA}
        \label{fig:tsne_voc_lora}
    \end{subfigure}
    \hspace{0.03\textwidth}
    \begin{subfigure}[t]{0.43\textwidth}
        \centering
        \tsneimg{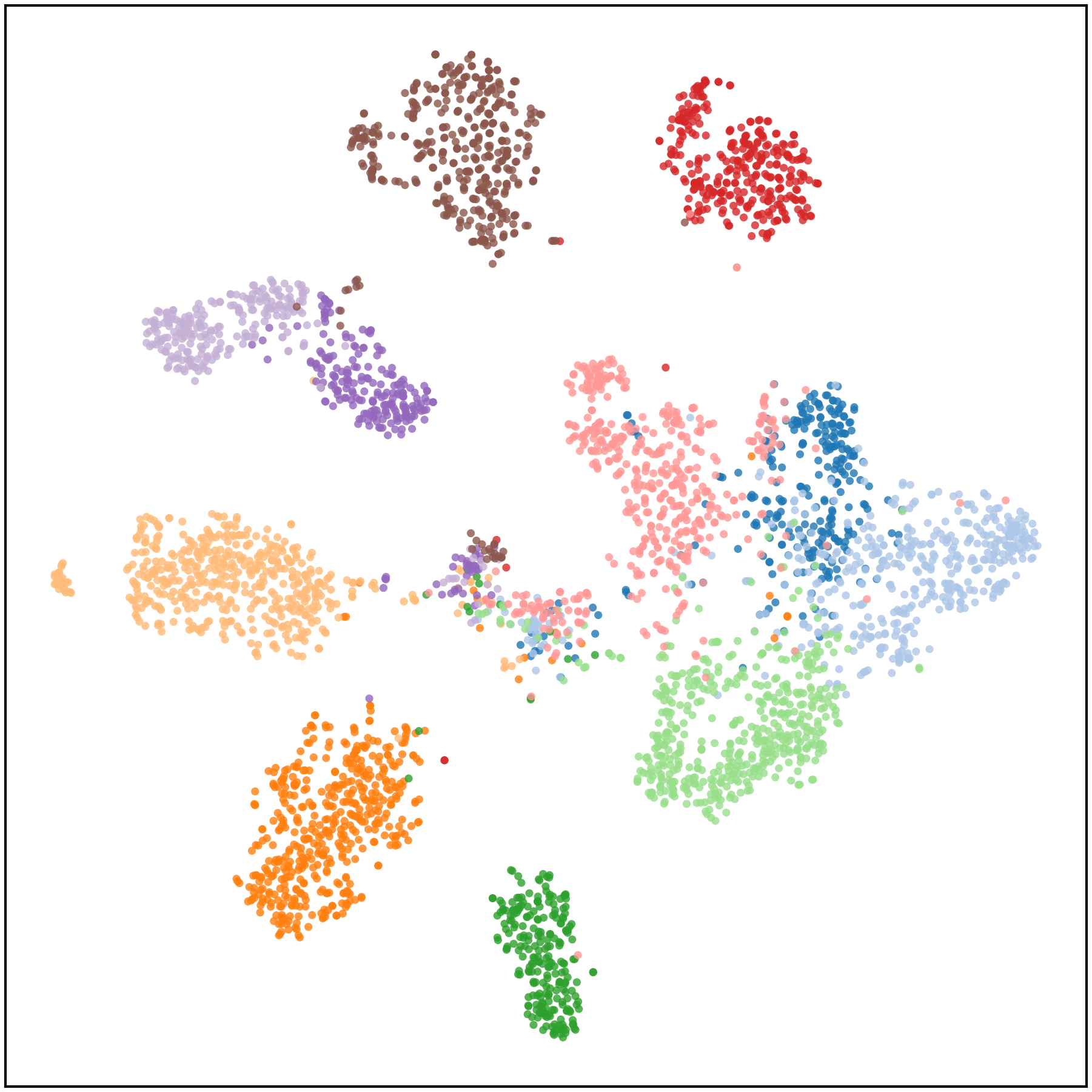}
        \caption{TT100K w/ LoRA}
        \label{fig:tsne_tt100k_lora}
    \end{subfigure}

    \caption{t-SNE visualization of category features on VOC and TT100K. Features are extracted from the original pretrained model and the LoRA-adapted model. Each color denotes a category.}
    \label{fig:tsne}
\end{figure}
\textbf{Plasticity for Classification Representations.}
Since the PTM maintains reliable localization under the frozen setting, its predicted boxes can be used to associate extracted features with object instances for analysis. We therefore use the same frozen setting as in the localization analysis to examine the discriminability of features extracted by the PTM.
Specifically, we match each ground-truth object to the prediction with the highest IoU and assign the corresponding ground-truth label to the matched feature.
Based on these features, we use t-SNE~\cite{van2008visualizing} to qualitatively visualize class separability.
Fig.~\ref{fig:tsne_voc_cls} and Fig.~\ref{fig:tsne_tt100k_cls} show the visualizations of features from VOC and TT100K, respectively.
On VOC, the features already form relatively separable clusters, suggesting that frozen pretrained representations provide adequate class discrimination in near in-domain settings.
However, this property does not necessarily transfer to more realistic cross-domain settings.
For example, on TT100K, the features exhibit substantial category overlap, suggesting that frozen pretrained representations are insufficient to establish clear decision boundaries.
To further examine whether additional plasticity improves classification-related representations, we insert LoRA modules into the Transformer layers of the PTM and visualize the adapted features.
As shown in Fig.~\ref{fig:tsne_voc_lora}, LoRA introduces only marginal changes in class separability on VOC, where the frozen features are already discriminative.
By contrast, LoRA produces clearly separated clusters on TT100K, as shown in Fig.~\ref{fig:tsne_tt100k_lora}.
The previously overlapping features become separable, with clearer boundaries emerging between categories.
These visualizations provide qualitative evidence that, compared with the generalizable localization ability of the pretrained detector, classification-related representations require additional plasticity for downstream adaptation.

\subsection{Selective Adaptation and Retention Framework}
\label{sec:decoupled}

The above analysis reveals a clear component-wise asymmetry in PTMIOD.
The localization branch provides generalizable localization priors and therefore benefits more from preservation, whereas classification-related components require sufficient plasticity to adapt to new categories while maintaining stability for previously learned decision boundaries.
This suggests that PTMIOD requires a selective adaptation strategy instead of indiscriminately updating the entire detector.

As shown in Sec.~\ref{sec:analysis}, the pretrained localization module remains robust under substantial domain shifts, suggesting that it primarily encodes domain-agnostic geometric priors rather than task-specific semantic cues.
Further updating this module offers limited benefit and may instead disrupt its pretrained localization capability.
Accordingly, we freeze the localization heads throughout all tasks.
In contrast, classification in PTMIOD requires stronger plasticity, especially when downstream tasks differ significantly from the pretraining domain.
Previous PTMIOD methods~\cite{an2025parameterized,bhatt2024preventing} mainly rely on prompt learning to elicit relevant knowledge from the PTM.\footnote{A more detailed discussion of prompt-based PTMIOD methods is provided in Appendix B.}
However, prompt tuning primarily exploits existing pretrained capabilities rather than adapting the representation space, and may therefore be insufficient when the target domain deviates markedly from the pretraining domain~\cite{petrov2023prompting}.
LoRA fine-tuning instead adapts the representation space more flexibly while causing limited disruption to pretrained knowledge, making it better suited to realistic scenarios.
Motivated by this observation, we apply LoRA to all transformer layers in the encoder $\mathcal{E}$ and decoder $\mathcal{D}$.
Specifically, we inject trainable low-rank components into the original weights $W \in \mathbb{R}^{d \times k}$:
\begin{equation}
W' = W + \Delta W = W + BA, \quad B \in \mathbb{R}^{d \times r}, A \in \mathbb{R}^{r \times k}
\end{equation}
where $r$ denotes the LoRA rank.
During incremental training, only the LoRA parameters and the classification heads \(\mathcal{F}^1\) and \(\mathcal{F}^2\) are updated.
The pretrained \(\mathcal{F}^1\) serves as a strong initialization for proposal selection; jointly updating it with our subsequent methods further adapts the selection process to new tasks while mitigating forgetting.

\subsection{Quality-aware Gaussian Feature Replay}
After decoupling localization preservation from classification adaptation, the remaining challenge lies in retaining old-class decision boundaries in the classification space.
A common strategy to alleviate forgetting is to replay old knowledge during subsequent tasks.
Existing replay-based methods typically use pseudo-labeling~\cite{gupta2022ow,kim2024vlm,liu2023continual} or exemplar replay~\cite{yuyang2023augmented,liu2023continual}.
However, pseudo-labeling becomes ineffective when old-category objects are absent from future-task data, while exemplar replay requires storing raw samples from previous tasks, which introduces additional storage overhead and potential privacy concerns.
To address these limitations, we introduce QGFR, a pseudo-feature replay mechanism tailored to DETR-style detectors that replaces raw exemplars with compact Gaussian distributions.
To make the estimated Gaussian distributions more representative, QGFR further adopts a quality-aware feature selection strategy that retains reliable features for distribution estimation.
Specifically, for each previously learned category, QGFR estimates and maintains a Gaussian distribution from the high-quality features, and then trains the classification heads with both synthetic features sampled from these distributions and real features from the current task.

\subsubsection{Quality-aware Gaussian Modeling}
\label{sec:feature_collection}
Obtaining high-quality features is critical for constructing effective pseudo-feature replay, especially in IOD.
However, in object detection, a model produces a large set of predictions, most of which are low-quality due to inaccurate localization or unreliable classification.
Using the low-quality features associated with these predictions to construct Gaussian distributions would introduce noise and degrade Gaussian modeling.
To mitigate this issue and obtain reliable distributions, QGFR performs quality-aware feature selection throughout the Gaussian modeling.
For each image, it first uses Hungarian matching to retain the query features that best match the ground-truth objects, forming a candidate pool.
It then refines this pool by filtering out features with low localization or classification quality, and fits a Gaussian distribution for each newly introduced category using the retained features.

Specifically, after learning task \(t\), we perform inference on \(D_t\) using \(\mathcal{M}_t\).
We first describe the decoder-stage feature collection illustrated in the left panel of Fig.~\ref{fig:framework}, omitting stage superscripts for clarity.
For each image, given the ground-truth objects $\{(y_i,\mathbf{b}_i)\}_{i=1}^{n}$ and the predictions $\{(\hat{\mathbf{s}}_j,\hat{\mathbf{b}}_j)\}_{j=1}^{k}$, where $k$ is much larger than $n$, we pad the ground-truth set with no-object entries $\varnothing$ to obtain a set of cardinality $k$.
To find the best bipartite matching between these two sets, we search for a permutation of $k$
elements $\sigma \in\mathfrak{S}_k$ with the lowest cost:
\begin{equation}
\hat{\sigma}
=
\arg\min_{\sigma \in \mathfrak{S}_k}
\sum_{i=1}^{k}
\mathcal{L}_{\mathrm{match}}\big((y_i,\mathbf{b}_i),(\hat{\mathbf{s}}_{\sigma(i)},\hat{\mathbf{b}
}_{\sigma(i)})\big),
\end{equation}
where $\sigma(i)$ is the index of the prediction matched to the $i$-th ground-truth object, and $\mathcal{L}_{\mathrm{match}}$ follows the matching cost of DETR.
Based on permutation \(\hat{\sigma}\), we retain the decoder features matched to the \(n\) ground-truth objects, denoted by \(\{\mathbf{z}_{\hat{\sigma}(i)}\}_{i=1}^{n}\), together with their corresponding labels \(\{y_i\}_{i=1}^{n}\).
To measure whether a collected feature is reliable for both classification and localization, we assign it a quality score by jointly considering its class confidence and spatial alignment.
For the feature \(\mathbf{z}_{\hat{\sigma}(i)}\), the score is defined as
\begin{equation}
q_i = \hat{s}_{\hat{\sigma}(i),\, y_i} \cdot \mathrm{IoU}(\hat{\mathbf{b}}_{\hat{\sigma}(i)}, \mathbf{b}_i),
\end{equation}
where \(\hat{s}_{\hat{\sigma}(i),\, y_i}\) is the classification score for $y_i$, and \(\mathrm{IoU}\) measures the spatial alignment between the predicted box and the corresponding ground-truth box.
This design assigns a high score only to features that are both semantically reliable and spatially accurate.
Aggregating all matched features over \(D_t\), we obtain the class-wise candidate feature pool:
\begin{equation}
\mathcal{P}_{t,c} =
\{(\mathbf{z}_{\hat{\sigma}(i)}, q_i) \mid y_i = c\}.
\end{equation}

\begin{figure}[t]
    \centering
    \includegraphics[width=\textwidth]{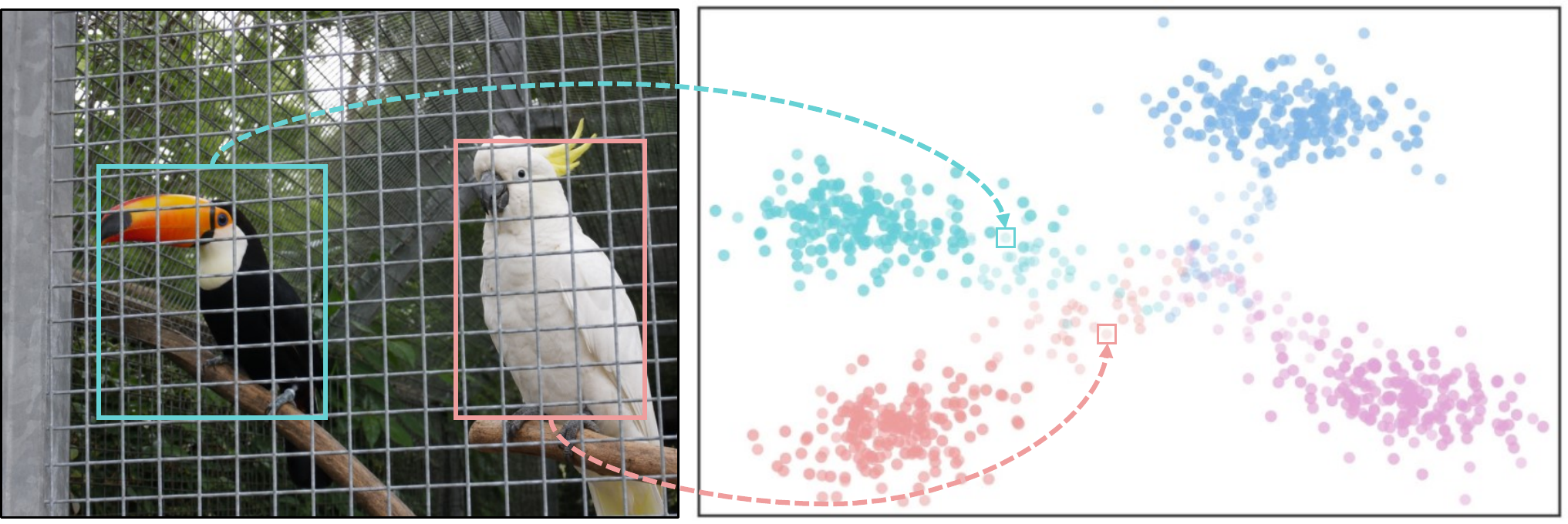}
    \caption{\textbf{Visualization of ambiguous object regions and collected features.} The left panel shows selected object regions affected by occlusion and background clutter. The right panel presents the t-SNE projection of collected class-aligned features from four categories with quality-aware coloring. Lighter points indicate lower quality scores.}
    \label{fig:feature_quality}
    
\end{figure}

\begin{figure}[t]
    \centering
    \includegraphics[width=\textwidth]{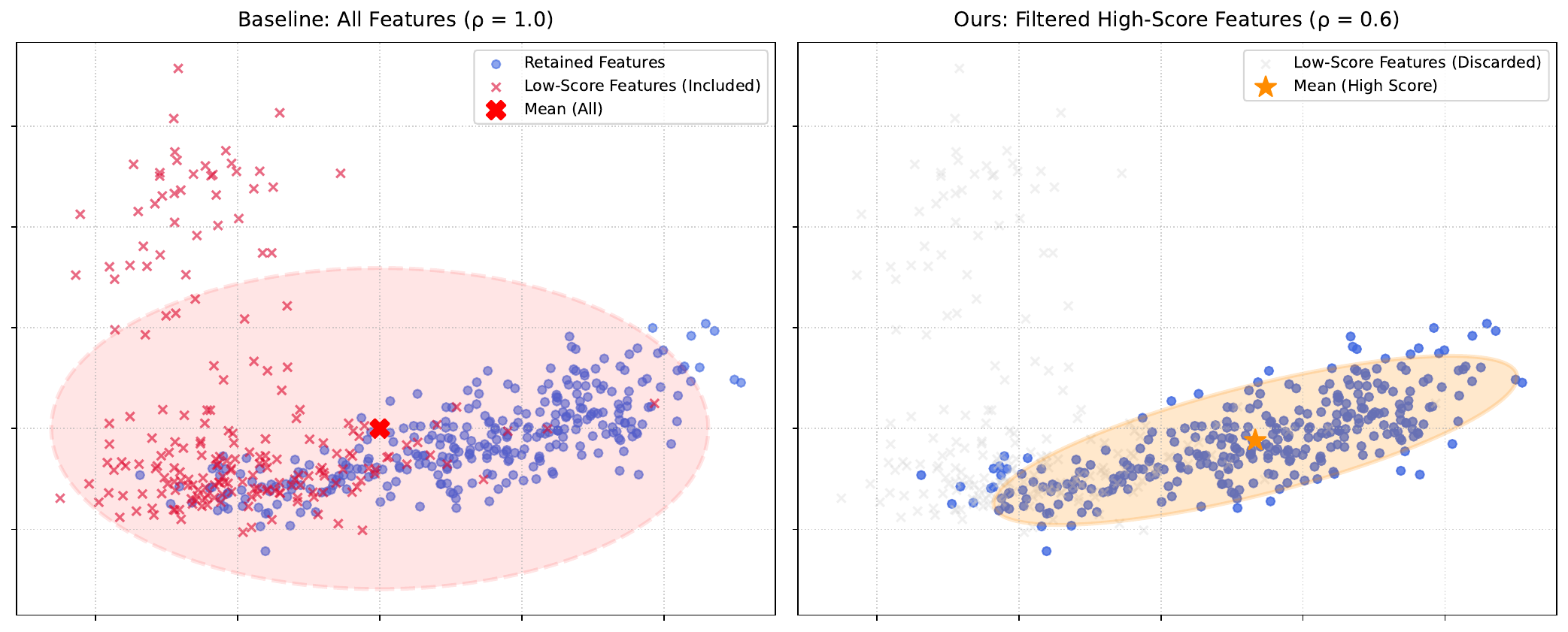}
    \caption{\textbf{Comparison of Gaussian modeling with and without low-quality feature filtering.} 
    For visualization, the collected features are projected into 2D by PCA \cite{shlens2014tutorial}. 
    \textbf{Left}: using all collected features to fit the Gaussian. 
    \textbf{Right}: filtering out low-score features before fitting.}
    \label{fig:high_quality_feature}
\end{figure}

As shown in Fig.~\ref{fig:feature_quality}, the matched instances can be ambiguous due to occlusion, background clutter, or low resolution. 
Such ambiguous features tend to have lower quality scores \(q_i\), and the t-SNE visualization shows that they tend to lie near class boundaries or overlap with features from other categories.
If used indiscriminately in Gaussian fitting, they bias the class mean and inflate the covariance, making the replayed pseudo-features less representative of old categories. 
To obtain reliable distributions, we adopt a quality-aware feature selection strategy before Gaussian fitting.
For each category \(c\), given the feature pool \(\mathcal{P}_{t,c}\), we sort all feature-score pairs by \(q_i\) in descending order and select the top \(\rho\) fraction of features with the highest quality scores.
Let \(\widetilde{\mathcal{P}}_{t,c} \subset \mathcal{P}_{t,c}\) denote the retained subset. 
We then fit a Gaussian distribution to the selected feature vectors in \(\widetilde{\mathcal{P}}_{t,c}\).\footnote{A detailed justification for the use of Gaussian fitting is provided in Appendix C.}
Specifically, for each category \(c\), we estimate the mean \(\boldsymbol{\mu}_{t,c}\in \mathbb{R}^{d}\) and covariance \(\boldsymbol{\Sigma}_{t,c}\in \mathbb{R}^{d\times d}\) from these feature vectors. The resulting replay distribution is defined as
\(\mathcal{N}(\boldsymbol{\mu}_{t,c}, \boldsymbol{\Sigma}_{t,c})\).
The encoder stage is handled in the same way by replacing  decoder features \(\mathbf{Z}^2\) with the encoder features \(\mathbf{Z}^1\), yielding an encoder-stage Gaussian distribution for each category.
In our implementation, storing the resulting Gaussian statistics requires only 1.13\,MB per category.
The effect of quality-aware selection strategy is qualitatively illustrated in Fig.~\ref{fig:high_quality_feature}.
Compared with fitting all collected features, filtering out low-quality features yields a more compact Gaussian distribution with a less biased mean and a tighter distribution boundary, making the replayed pseudo-features more representative of previously learned categories.

\subsubsection{ Replay with High Quality Feature} 
To mitigate catastrophic forgetting in the classification heads, we jointly train the classification heads using real features from the current task and pseudo-features sampled from old-class Gaussian distributions during incremental learning of task \(t\) (\(t>1\)). 
Specifically, the pseudo-features are sampled as
\begin{equation}
\tilde{\mathbf{z}}_{i,c}^* \sim \mathcal{N}(\boldsymbol{\mu}_c^*, \boldsymbol{\Sigma}_c^*), \quad c \in C_{1:t-1}, \quad * \in \{1,2\},
\end{equation}
where \(\boldsymbol{\mu}_c^*\) and \(\boldsymbol{\Sigma}_c^*\) are the cached Gaussian mean and covariance of previously learned class \(c\) at stage \(*\).
Before each epoch, we sample $\frac{N_t}{|C_{1:t-1}|}$ pseudo-features for each old class per stage, where $N_t$ is the number of annotated instances in task $t$. 
We then distribute these pseudo-features uniformly across training iterations, replaying $\frac{N_t}{S_t}$ pseudo-features per iteration, where $S_t$ is the number of iterations per epoch.
Thus, each epoch replays the same total number of pseudo-features as annotated instances in the current task, which balances replay from old classes with learning from the current task.
For diversity, we resample pseudo-features at the beginning of each epoch.

The sampled pseudo-features are then used to supervise the classification heads at both detection stages.
Specifically, we define the labeled pseudo-feature sets as
\begin{equation}
\tilde{\mathcal{Z}}^1 =
\{(\tilde{\mathbf{z}}_{i,c}^{1}, c) \mid c \in C_{1:t-1}\},
\quad
\tilde{\mathcal{Z}}^2 =
\{(\tilde{\mathbf{z}}_{i,c}^{2}, c) \mid c \in C_{1:t-1}\}.
\end{equation}
The features in \(\tilde{\mathcal{Z}}^1\) are passed to the proposal classifier \(\mathcal{F}^1\) for proposal-level retention, while those in \(\tilde{\mathcal{Z}}^2\) are passed to the final classifier \(\mathcal{F}^2\) for classification-level retention.
Because the replayed features are directly passed to the classification heads, the additional computational cost is low.
The QGFR losses are defined as:

\begin{equation}
\mathcal{L}_{\text{rpy}} =\mathcal{L}_{\text{rpy}}^1 + \mathcal{L}_{\text{rpy}}^2 ,
\label{eq:tol_loss}
\end{equation}
where \(\mathcal{L}_{\text{rpy}}^1\) and \(\mathcal{L}_{\text{rpy}}^2\) are computed as:

\begin{equation}
\begin{aligned}
\mathcal{L}_{\text{rpy}}^1
&= -\sum_{(\tilde{\mathbf{z}}_{i,c}^{1}, c) \in \tilde{\mathcal{Z}}^1}
\log\big(\mathcal{F}^1(\tilde{\mathbf{z}}_{i,c}^{1})_c\big), \\
\mathcal{L}_{\text{rpy}}^2
&= -\sum_{(\tilde{\mathbf{z}}_{i,c}^{2}, c) \in \tilde{\mathcal{Z}}^2}
\log\big(\mathcal{F}^2(\tilde{\mathbf{z}}_{i,c}^{2})_c\big),
\end{aligned}
\end{equation}
where \(\mathcal{F}^*(\tilde{\mathbf{z}}_{i,c}^{*})_c\) denotes the predicted probability of class \(c\).

\subsection{Stabilizing Feature Space via Two-stage Consistent Distillation}
\label{sec:tcd}
QGFR requires stable feature representations for previously learned classes; otherwise, the sampled pseudo-features may become misaligned with the current feature space and provide unreliable supervision for old-class retention.
A simple solution is to freeze the LoRA parameters after the initial task, which substantially constrains the plasticity required for subsequent task adaptation.
An alternative is to continue adapting the model while constraining its representations through knowledge distillation (KD).
However, directly applying standard KD to two-stage DETR detectors is suboptimal, because the teacher and student are not naturally aligned throughout the two-stage process.
Specifically, they may select different Top-\(k\) spatial locations in the encoder stage and different proposals in the decoder stage, which leads to inconsistent distillation targets.
To address this issue, we propose \textbf{Two-stage Consistent Distillation (TCD)}. 
As illustrated in the right panel of Fig.~\ref{fig:framework}, TCD enforces consistency between the teacher and the student at both stages of DINO by distilling encoder features at the same Top-\(k\) indices and decoder features conditioned on the same proposals generated by the teacher and the same encoder memory.

For task \(t\), the teacher is initialized from \(\mathcal{M}_{t-1}\).
At the encoder stage, TCD uses the teacher’s classification scores to select the Top-\(k\) locations for feature alignment:
\begin{equation}
\mathcal{I}_{\mathrm{tc}} = \mathrm{Top}\text{-}k\big(\{\max_c s_{i,c}^{1,\mathrm{tc}}\}\big),
\end{equation}
where \(s_{i,c}^{1,\mathrm{tc}}\) denotes the teacher’s classification score for class \(c\) at location \(i\), and \(\mathcal{I}_{\mathrm{tc}}\) is the resulting set of selected locations.
The corresponding student and teacher features are then gathered as
\begin{equation}
\mathbf{Z}_{\mathcal{I}_{\mathrm{tc}}}^{1,\mathrm{stu}} = [\mathbf{z}_{i}^{1,\mathrm{stu}}]_{i\in \mathcal{I}_{\mathrm{tc}}}, 
\qquad
\mathbf{Z}_{\mathcal{I}_{\mathrm{tc}}}^{1,\mathrm{tc}} = [\mathbf{z}_{i}^{1,\mathrm{tc}}]_{i\in \mathcal{I}_{\mathrm{tc}}}.
\end{equation}
Because both feature sets are gathered using the same locations selected by the teacher, TCD provides spatially aligned supervision and avoids mismatches that could arise from separate location selection by the teacher and student encoders.
The consistency loss at the encoder stage is defined as
\begin{equation}
\mathcal{L}_{\mathrm{tcd}}^{1}
=
\left\|
\mathbf{Z}_{\mathcal{I}_{\mathrm{tc}}}^{1,\mathrm{stu}}
-
\mathbf{Z}_{\mathcal{I}_{\mathrm{tc}}}^{1,\mathrm{tc}}
\right\|_2^2.
\end{equation}

TCD further enforces consistency at the decoder stage. In two-stage DINO, decoder refinement relies on encoder memory \(\mathbf{Z}^1\) and proposal anchors \(Q^\mathrm{pos}\), both derived from the encoder.
Therefore, direct distillation can become unreliable when the teacher and student are conditioned on different encoder features or different proposals. 
To eliminate this mismatch, we construct a consistent distillation path by providing both decoders with the same inputs from the teacher. 
Specifically, we use the teacher encoder memory together with the Top-\(k\) proposals generated by the teacher in the first stage. These proposals serve as decoder anchors:
\begin{equation}
Q_{\mathrm{tc}}^{\mathrm{pos}}
=
\{\mathbf{\hat{b}}_{i,\mathrm{tc}}^1 \mid i\in \mathcal{I}_{\mathrm{tc}}\}.
\end{equation}
The shared inputs \((\mathbf{Z}_{\mathrm{tc}}^1,Q_{\mathrm{tc}}^\mathrm{pos})\) are then fed into both the teacher and student decoders.
This path is used only for distillation, while the student's main detection branch still operates on its own encoder memory and proposals. 
Let \(\mathcal{D}_\mathrm{stu}^l(\cdot,\cdot,\cdot)\) and \(\mathcal{D}_{\mathrm{tc}}^l(\cdot,\cdot,\cdot)\) denote the outputs of the \(l\)-th decoder layer of the student and teacher, respectively. The consistency loss at the decoder stage is defined as
\begin{equation}
\mathcal{L}_{\mathrm{tcd}}^{\mathrm{2}}
=
\sum_{l=1}^{L}
\left\|
\mathcal{D}_\mathrm{stu}^l(\mathbf{Z}_{\mathrm{tc}}^1, Q_{\mathrm{tc}}^{\mathrm{pos}},Q^{\mathrm{con}})
-
\mathcal{D}_{\mathrm{tc}}^l(\mathbf{Z}_{\mathrm{tc}}^1, Q_{\mathrm{tc}}^{\mathrm{pos}},Q^{\mathrm{con}})
\right\|_2^2,
\end{equation}
where \(L\) is the number of decoder layers. The final TCD loss is given by
\begin{equation}
\mathcal{L}_{\mathrm{tcd}}
=
\mathcal{L}_{\mathrm{tcd}}^{\mathrm{1}}
+
\mathcal{L}_{\mathrm{tcd}}^{\mathrm{2}}.
\end{equation}

\subsection{Total Objectives}
For the first task (\(t=1\)), no old-class knowledge is available for replay or distillation.
Therefore, the training objective reduces to the standard DINO detection loss:
\begin{equation}
\mathcal{L}_{\mathrm{total}} = \mathcal{L}_{\mathrm{DINO}}, \qquad t=1.
\end{equation}

For subsequent incremental tasks (\(t>1\)), the detector is optimized by combining the original DINO loss with the QGFR loss and the proposed Two-stage Consistent Distillation loss:
\begin{equation}
\mathcal{L}_{\mathrm{total}}
=
\mathcal{L}_{\mathrm{DINO}}
+
\lambda_1 \mathcal{L}_{\mathrm{rpy}}
+
\lambda_2 \mathcal{L}_{\mathrm{tcd}},
\qquad t>1,
\end{equation}
where  \(\lambda_1\) and \(\lambda_2\) are balancing coefficients that control the contributions of replay and distillation, respectively.

\section{Experiments}
\subsection{Setup}
\textbf{Datasets and Evaluation Metric.} 
We conduct experiments on three datasets: MS COCO 2017~\cite{lin2014microsoft}, Pascal VOC 2007~\cite{everingham2010pascal}, and TT100K~\cite{zhu2016traffic}.
COCO and VOC are used as in-domain tasks, with 50\% and 38.9\% overlap with the pretraining dataset, Objects365~\cite{shao2019objects365}, respectively. 
Both datasets are commonly used for IOD evaluation: COCO is larger in scale and contains 80 categories, while VOC is smaller and contains 20 categories.
In contrast, TT100K is included to assess the robustness of our framework in a more challenging cross-domain scenario. We use a subset of TT100K containing the 20 categories with the highest instance counts, which has only 0.217\% overlap with Objects365.
Here, \textit{overlap} is defined as the ratio of instances in the pretraining dataset whose categories belong to the downstream task to all instances in the pretraining dataset.
We adopt mean Average Precision (mAP) as the evaluation metric.
For COCO, we report mAP across IoU thresholds from 0.50 to 0.95 (mAP$_{50:95}$), as well as mAP at 0.50 (mAP$_{50}$) and 0.75 (mAP$_{75}$) to provide a comprehensive assessment.
For VOC and TT100K, we report mAP$_{50}$ as the primary metric.

\textbf{Baselines.}
We compare our framework with classical and recent incremental object detection baselines, including Faster R-CNN-based, DETR-based, and PTM-based approaches.
Specifically, we compare with Faster ILOD~\cite{peng2020faster}, MMA~\cite{cermelli2022modeling}, ABR~\cite{yuyang2023augmented}, CL-DETR~\cite{liu2023continual}, MD-DETR~\cite{bhatt2024preventing}, and P$^2$IOD~\cite{an2025parameterized}.
Among these baselines, MD-DETR and P$^2$IOD are the most relevant PTMIOD competitors.
For a fair comparison, we reproduce these methods on DINO with the same pretrained initialization when official results are unavailable.

\textbf{IOD Setting.} 
Incremental object detection is typically evaluated under two common settings: single-increment (consisting of two tasks) and multi-increment (involving more than two tasks).
For COCO, we adopt two single-increment settings, 40-40 and 70-10.
For VOC and TT100K, we use the same set of configurations.
Specifically, we adopt one single-increment setting, 15-5, and two multi-increment settings, 10-5 and 15-1, where the second number denotes the number of new classes introduced at each incremental step.
For each setting, we report mAP$_{50}$ after the final incremental step on the \textit{base classes} (e.g., classes 1-10 in the 10-5 setting), the \textit{incremental classes} (e.g., classes 11-20 in the 10-5 setting), and \textit{all classes} (i.e., classes 1-20 in all settings).
Further details on the experimental setup, including the multi-increment setting on COCO, are provided in Appendix D.

\subsection{Main Results}
\begin{table*}[t]
\centering
\small
\caption{mAP$_{50}$ on VOC and TT100K. Best results are in \textbf{bold}, and second-best results are \underline{underlined}. All TT100K results are reproduced with our implementation. On VOC, results marked with $^\dagger$ are reproduced with our implementation, and the remaining results are taken from official reports or~\cite{an2025parameterized}.}
\label{tab:tt_voc}
\setlength{\tabcolsep}{3.5pt}
\renewcommand{\arraystretch}{1.12}
\resizebox{\textwidth}{!}{%
\begin{tabular}{l|ccc|ccc|ccc||ccc|ccc|ccc}
\toprule
\multirow{3}{*}{\textbf{Method}} 
& \multicolumn{9}{c||}{\textbf{VOC}} 
& \multicolumn{9}{c}{\textbf{TT100K}} \\
\cmidrule(lr){2-10} \cmidrule(lr){11-19}
& \multicolumn{3}{c|}{\textbf{15--5 (2 Tasks)}} 
& \multicolumn{3}{c|}{\textbf{10--5 (3 Tasks)}} 
& \multicolumn{3}{c||}{\textbf{15--1 (6 Tasks)}} 
& \multicolumn{3}{c|}{\textbf{15--5 (2 Tasks)}} 
& \multicolumn{3}{c|}{\textbf{10--5 (3 Tasks)}} 
& \multicolumn{3}{c}{\textbf{15--1 (6 Tasks)}} \\
\cmidrule(lr){2-4} \cmidrule(lr){5-7} \cmidrule(lr){8-10}
\cmidrule(lr){11-13} \cmidrule(lr){14-16} \cmidrule(lr){17-19}
& \textbf{Base} & \textbf{Inc} & \textbf{All}
& \textbf{Base} & \textbf{Inc} & \textbf{All}
& \textbf{Base} & \textbf{Inc} & \textbf{All}
& \textbf{Base} & \textbf{Inc} & \textbf{All}
& \textbf{Base} & \textbf{Inc} & \textbf{All}
& \textbf{Base} & \textbf{Inc} & \textbf{All} \\
\midrule
Faster ILOD~\cite{peng2020faster}       & 73.1 & 57.3 & 69.2 & 68.3 & 57.9 & 63.1 & 66.9 & 44.5 & 61.3 & 46.7 & 27.3 & 41.9 & 45.6 & 19.5 & 32.6 & \underline{47.2} & 24.2 & \underline{41.4} \\
MMA~\cite{cermelli2022modeling}               & 72.7 & 60.6 & 69.7 & 67.4 & 60.5 & 64.0 & 67.2 & 47.8 & 62.3 & 44.4 & 35.4 & 42.2 & 50.8 & 26.3 & 38.6 & 43.5 & 29.0 & 39.9 \\
ABR~\cite{yuyang2023augmented}               & 73.0 & 65.1 & 71.0 & 68.7 & 67.1 & 67.9 & 68.7 & 56.7 & 65.7 & \underline{47.3} & 39.0 & \underline{45.6} & \underline{52.5} & \underline{33.8} & \underline{43.1} & 43.1 & \underline{35.4} & 41.0 \\
CL-DETR~\cite{liu2023continual} 
& 52.1$^\dagger$ & 38.2$^\dagger$ & 48.6$^\dagger$ 
& 15.0$^\dagger$ & 26.1$^\dagger$ & 20.6$^\dagger$ 
& 45.3$^\dagger$ & 10.1$^\dagger$ & 36.5$^\dagger$ 
& 36.8 & 43.9 & 38.6 
& 49.8 & 28.0 & 38.9 
& 30.8 & 9.4 & 25.4 \\
\midrule
MD-DETR~\cite{bhatt2024preventing} & 86.1 & 84.7 & 85.8 & 80.1 & 87.5 & 83.8 & 77.5$^\dagger$ & 49.7$^\dagger$ & 70.5$^\dagger$ & 21.2 & 47.1 & 27.7 & 19.5 & 27.7 & 23.6 & 11.7 & 18.0 & 13.3 \\
P$^2$IOD~\cite{an2025parameterized} 
& \underline{91.2} & \underline{85.2} & \underline{89.7} 
& \underline{89.1} & \underline{89.1} & \underline{89.1} 
& \underline{89.4}$^\dagger$ & \underline{81.1}$^\dagger$ & \underline{87.4}$^\dagger$ 
& 33.1 & \underline{48.8} & 37.0 
& 25.8 & 32.8 & 29.3
& 12.9 & 18.3 & 14.3 \\
\rowcolor{gray!8}
Ours              & \textbf{94.1} & \textbf{90.1} & \textbf{93.1} & \textbf{91.9} & \textbf{90.7} & \textbf{91.3} & \textbf{91.2} & \textbf{86.9} & \textbf{90.1} & \textbf{84.7} & \textbf{65.3} & \textbf{79.9} & \textbf{78.2} & \textbf{45.3} & \textbf{61.7} & \textbf{80.1} & \textbf{54.3} & \textbf{73.6} \\
\bottomrule
\end{tabular}%
}
\end{table*}
Table~\ref{tab:tt_voc} reports results on VOC and TT100K under the same incremental configurations, covering in-domain and cross-domain scenarios, respectively. Table~\ref{tab:result-coco} further reports results on the more challenging in-domain COCO benchmark.
On VOC, the proposed framework consistently achieves the best results across all incremental settings. Compared with the strongest baseline P$^2$IOD, it improves the \textit{all classes} mAP$_{50}$ by 3.4, 2.2, and 2.7 points under the 15-5, 10-5, and 15-1 settings, respectively. 
Notably, it also achieves the best mAP$_{50}$ on both \textit{base classes} and \textit{incremental classes} across all three settings, indicating that the proposed framework substantially improves plasticity while preserving stability.
The advantage becomes more pronounced on TT100K, where the cross-domain setting poses a more challenging and realistic scenario. In this setting, our framework outperforms the strongest baseline in terms of mAP$_{50}$ on \textit{all classes} by 34.3, 18.6, and 32.2 points under the 15-5, 10-5, and 15-1 settings, respectively.
Compared with existing PTMIOD methods, these results demonstrate that our method not only performs well on standard in-domain incremental detection, but also generalizes effectively to cross-domain incremental scenarios.
On COCO,  our framework consistently outperforms existing approaches under both the 40–40 and 70–10 settings across all IoU thresholds. Compared with P$^2$IOD, it improves mAP$_{50:90}$ by 3.4 and 4.2 points, respectively, indicating that the proposed framework achieves a better balance between stability and plasticity in more challenging in-domain scenarios.
\begin{table}[t]
\centering
\footnotesize
\caption{mAP results on COCO under different IoU thresholds. Results of competing methods are taken from~\cite{an2025parameterized}. Best results are in \textbf{bold}, and second-best results are \underline{underlined}.}
\label{tab:result-coco}
\setlength{\tabcolsep}{5.5pt}
\renewcommand{\arraystretch}{1.12}
\begin{tabular}{l|ccc|ccc}
\toprule
\multirow{2}{*}{\textbf{Method}} 
& \multicolumn{3}{c|}{\textbf{40-40}} 
& \multicolumn{3}{c}{\textbf{70-10}} \\
\cmidrule(lr){2-4} \cmidrule(lr){5-7}
& \textbf{mAP$_{50:95}$} & \textbf{mAP$_{50}$} & \textbf{mAP$_{75}$} 
& \textbf{mAP$_{50:95}$} & \textbf{mAP$_{50}$} & \textbf{mAP$_{75}$} \\
\midrule
Faster ILOD~\cite{peng2020faster}          & 20.6 & 40.1 & - & 21.3 & 39.9 & - \\
MMA~\cite{cermelli2022modeling}          & 33.0 & 56.6 & 34.6 & 30.2 & 52.1 & 31.5 \\
ABR~\cite{yuyang2023augmented}        & 34.5 & 57.8 & 35.2 & 31.1 & 52.9 & 32.7 \\
CL-DETR~\cite{liu2023continual}      & 42.0 & 60.1 & 45.9 & 40.4 & 58.0 & 43.9 \\
\midrule
MD-DETR~\cite{bhatt2024preventing}  & 50.0 &65.7 & 55.1 & 50.8 & 65.6 & 55.8 \\
P$^2$IOD~\cite{an2025parameterized}  & \underline{54.7} & \underline{71.1} & \underline{60.3} & \underline{55.2} & \underline{71.9} & \underline{60.7} \\
\rowcolor{gray!8}
Ours               & \textbf{58.1} & \textbf{74.6} & \textbf{64.2} & \textbf{59.4} & \textbf{75.9} & \textbf{65.5} \\
\bottomrule
\end{tabular}
\end{table}

\subsection{Ablation Studies}
We conduct ablation studies to systematically validate the proposed framework from multiple perspectives.
We first examine the effect of freezing pretrained localization heads during incremental learning.
With the localization heads frozen, we then ablate key framework factors to quantify the contribution of LoRA adaptation and the proposed loss terms.
Next, we compare QGFD with pseudo-labeling under old-class absence, where incremental data contain no old-class objects for pseudo-label generation.
We further analyze different distillation strategies to verify the effectiveness of the proposed TCD.
Finally, we study the sensitivity of important hyperparameters and their impact on the stability-plasticity trade-off.

\subsubsection{Effect of Freezing Pretrained Localization Heads}
To examine whether the pretrained localization heads should be updated during incremental learning, we compare two variants under the 15-1 setting.
Both variants keep the proposed framework unchanged except for the update strategy of the localization heads.
Table~\ref{tab:ablation_locfreeze} shows that freezing the localization heads consistently outperforms tuning them on both VOC and TT100K, improving the mAP on \textit{all classes} from 85.8 to 90.1 and from 65.7 to 73.6, respectively.
This verifies that freezing the pretrained localization heads is beneficial in both in-domain and cross-domain scenarios, because it preserves the strong generalization ability of the pretrained localization heads and avoids unnecessary forgetting caused by continual tuning.
\begin{table}[t]
\centering 
\caption{Effect of tuning versus freezing the localization heads under the 15--1 setting.}
\label{tab:ablation_locfreeze} 
\setlength{\tabcolsep}{3.5pt} 
\renewcommand{\arraystretch}{0.95} 
\small 
\begin{tabular}{lcccccc} 
\toprule 
\multirow{2}{*}{Variant} & \multicolumn{3}{c}{VOC} & \multicolumn{3}{c}{TT100K} \\ 
\cmidrule(lr){2-4} \cmidrule(lr){5-7} & Base & Inc & All & Base & Inc & All \\ 
\midrule Tune loc. heads & 86.9 & 82.5 & 85.8 & 71.4 & 48.8 & 65.7 \\ 
Freeze loc. heads & 91.2 & 86.9 & 90.1 & 80.1 & 54.3 & 73.6 \\ 
\bottomrule 
\end{tabular} 
\end{table}

\subsubsection{Ablation of Key Framework Factors}
\begin{table*}
\caption{Cumulative ablation study on the key factors of our framework on VOC and TT100K. 
In this table, the localization head is always kept frozen, and the FT baseline updates only the classification head and does not use LoRA or any other proposed designs.}
\label{tab:ablation}
\centering
\centering
\footnotesize
\resizebox{\textwidth}{!}{
\begin{tabular}{l|lll|lll||lll|lll} 
\hline
\multicolumn{1}{l|}{\textbf{Dataset}} & \multicolumn{6}{c||}{\textbf{VOC}} & \multicolumn{6}{c}{\textbf{TT100K}} \\ 
\hline
\textbf{Setting} & \multicolumn{3}{c|}{\textbf{15--5 (2 Tasks)}} & \multicolumn{3}{c||}{\textbf{15--1 (6 Tasks)}} & \multicolumn{3}{c|}{\textbf{15--5 (2 Tasks)}} & \multicolumn{3}{c}{\textbf{15--1 (6 Tasks)}} \\ 
\hline
\textbf{Method} & \textbf{Base} & \textbf{Inc} &  \textbf{All}  & \textbf{Base} & \textbf{Inc} &  \textbf{All} & \textbf{Base} & \textbf{Inc} &  \textbf{All} & \textbf{Base} & \textbf{Inc} &  \textbf{All} \\ 
\hline
FT & 69.5 & 88.4 &  74.3 & 27.7 & 60.1 &  35.8 & 0.0 & 52.4 &  13.1 & 0.0 & 8.9 &  2.2 \\
+LoRA  & 83.5 & 89.6 &  85.0 & 46.1 & 59.9 &  49.6 & 41.4 & 71.0 &  48.8 &  1.0 & 16.7 &  4.9  \\
++$\mathcal{L}_{\mathrm{rpy}}$  & 92.1 & 91.0 &  91.8 & 79.0 & 79.4 &  79.1 & 26.9 & 68.9 &  37.4 & 6.9 & 14.4 &  8.8 \\
+++$\mathcal{L}_{\mathrm{tcd}}$  & 94.1 & 90.1 &  \textbf{93.1} & 91.2 & 86.9 &  \textbf{90.1 }& 84.7 & 65.3 &  \textbf{79.9} & 80.1 & 54.3 &  \textbf{73.6} \\
\hline
\end{tabular}
}
\end{table*}

We conduct a cumulative ablation study to evaluate how each major factor contributes to the overall framework.
As shown in Table~\ref{tab:ablation}, the fine-tuning (FT) baseline without these factors struggles to retain old knowledge and learn new classes, especially in multi-step and cross-domain settings.
On VOC 15-1, the mAP on \textit{base classes} drops to 27.7, indicating severe forgetting across multiple incremental steps. 
On TT100K, FT performs poorly on \textit{incremental classes}, indicating its limited ability to adapt to the cross-domain setting.
Introducing LoRA substantially improves the plasticity of the model, with particularly clear gains on TT100K. 
Under the TT100K 15-5 setting, LoRA improves the mAP on \textit{base classes} from 0.0 to 41.4 and the mAP on \textit{incremental classes} from 52.4 to 71.0. 
The improvement on \textit{base classes} suggests that LoRA enables stronger adaptation to the initial cross-domain task; thus, even after subsequent incremental learning, it retains much higher performance than FT. 
However, LoRA alone still provides limited protection against forgetting, especially under the multi-step 15-1 setting.
Adding $\mathcal{L}_{\mathrm{rpy}}$ substantially improves knowledge retention on VOC. 
Under the VOC 15-1 setting, the mAP on \textit{base classes} increases from 46.1 to 79.0, and the mAP on \textit{all classes} increases from 49.6 to 79.1. 
This shows that QGFD effectively preserves previously learned knowledge during incremental learning. 
However, $\mathcal{L}_{\mathrm{rpy}}$ alone becomes insufficient when the feature distribution shifts substantially, either across domains or across multiple incremental steps. 
For example, although it improves the mAP on \textit{all classes} on VOC 15-1, it remains far below the full model. 
Similarly, its performance on TT100K is still limited under both 15-5 and 15-1. 
This is because the Gaussian statistics estimated for old classes may become less reliable when the feature space drifts substantially, making replayed pseudo features less representative of the current model distribution and potentially weakening old knowledge preservation.
Introducing $\mathcal{L}_{\mathrm{tcd}}$ further addresses this issue by providing aligned distillation supervision between the teacher and student models. 
With $\mathcal{L}_{\mathrm{tcd}}$, the mAP on \textit{all classes} increases from 79.1 to 90.1 on VOC 15-1, from 37.4 to 79.9 on TT100K 15-5, and from 8.8 to 73.6 on TT100K 15-1. 
These results indicate that TCD is crucial for stabilizing QGFD when feature distributions shift significantly across domains or across multiple incremental steps.

\subsubsection{QGFD vs. Pseudo-Labeling under Old-Class Absence}
We compare pseudo-labeling with QGFD on VOC 10-5 to verify whether QGFD is more robust when future-task data contain no old-class objects for generating pseudo labels.
The result is shown in Table~\ref{tab:ablation_scene}.
We consider two scenarios that differ only in the incremental training data.
The standard scenario uses the original incremental training data, whereas the challenging scenario removes images containing old-class objects from the incremental training data, ensuring that no old-class objects appear during incremental training.
Both scenarios are evaluated on the same validation set, and all methods follow the same training protocol within each scenario.
A smaller performance drop from the standard scenario to the challenging scenario indicates stronger robustness to missing old-class objects and better preservation of old-class knowledge.
The mAP on \textit{base classes} drops by 9.8 and 2.4 points for MD-DETR and P$^2$IOD, respectively, under the challenging scenario.
These drops indicate that methods relying on pseudo-labeling are sensitive to the availability of old-class objects in the incremental training data, because old-class supervision becomes unavailable once such objects are removed.
In contrast, our framework shows only a marginal decrease of 0.2 points, from 91.9 to 91.7.
This stability suggests that QGFD does not depend on future-task images to recover old-class supervision; instead, it preserves old-class decision boundaries through the stored feature distributions.
Therefore, QGFD provides a more robust mechanism for old-class retention when new-task data contain no old-class objects.

\begin{table}[t]
\centering
\caption{Comparison on VOC 10--5 under standard and challenging scenarios. In the challenging scenario, no old-class objects appear in incremental training data. $\Delta$ denotes the mAP drop from the standard scenario to the challenging scenario.}
\label{tab:ablation_scene}
\setlength{\tabcolsep}{3.2pt}
\renewcommand{\arraystretch}{1.05}
\small
\begin{tabular}{lccc ccc cc}
\toprule
\multirow{2}{*}{Method}
& \multicolumn{3}{c}{Standard}
& \multicolumn{3}{c}{Challenging}
& \multicolumn{2}{c}{$\Delta$ $\downarrow$} \\
\cmidrule(lr){2-4} \cmidrule(lr){5-7} \cmidrule(lr){8-9}
& Base & Inc & All
& Base & Inc & All
& Base & All \\
\midrule
MD-DETR
& 80.1 & 87.5 & 83.8
& 70.3 & 79.4 & 74.8
& 9.8 & 9.0 \\

P$^2$IOD
& 89.1 & 89.1 & 89.1
& 86.7 & 81.6 & 84.1
& 2.4 & 5.0 \\

Ours
& 91.9 & 90.7 & \textbf{91.3}
& 91.7 & 88.9 & \textbf{90.7}
& \textbf{0.2} & \textbf{0.6} \\

\bottomrule
\end{tabular}
\end{table}

\subsubsection{Comparison of Different Distillation Strategies}
\begin{table}[t]
\centering
\caption{Effect of different distillation strategies under the 15--5 setting.}
\label{tab:ablation_distill}
\setlength{\tabcolsep}{3.5pt}
\renewcommand{\arraystretch}{0.95}
\small
\begin{tabular}{lcccccc}
\toprule
\multirow{2}{*}{Variant} & \multicolumn{3}{c}{VOC} & \multicolumn{3}{c}{TT100K} \\
\cmidrule(lr){2-4} \cmidrule(lr){5-7}
& Base & Inc & All & Base & Inc & All \\
\midrule
No KD       & 92.1 & 91.0 & 91.8 & 26.9 & 68.9 & 37.4 \\
Standard KD & 88.7 & 87.0 & 88.3 & 36.4 & 51.0 & 40.1 \\
TCD         & 94.1 & 90.1 & \textbf{93.1} & 84.7 & 65.3 & \textbf{79.9} \\
\bottomrule
\end{tabular}
\end{table}
\begin{figure}[t]
    \centering
    \includegraphics[width=\textwidth]{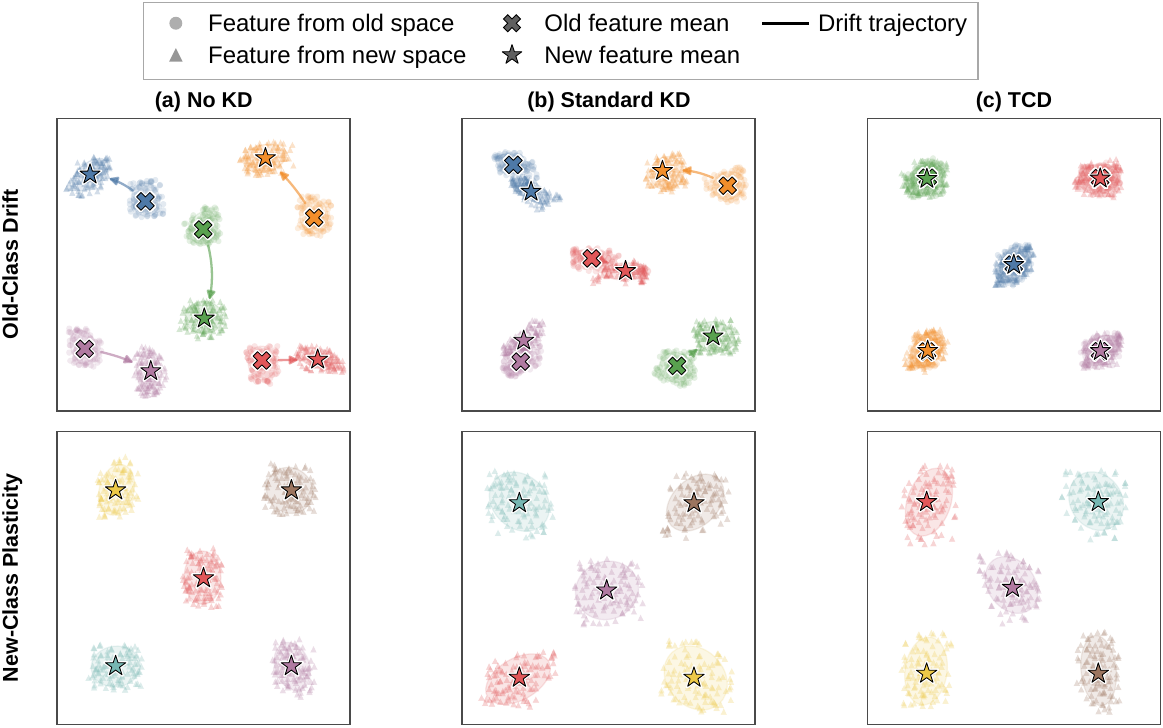}
    \caption{\textbf{Comparison of feature-space evolution under different distillation strategies on TT100K under the 15-5 setting.}
    The first row visualizes the drift of old-class features caused by the shift of the feature space after incremental learning, while the second row shows the clustering quality of newly learned classes in the updated feature space.
    From left to right, the three columns correspond to no distillation, standard knowledge distillation, and the proposed Two-stage Consistent Distillation (TCD), respectively.}
    \label{fig:tsne_compare_all_2x3}
    
\end{figure}
We compare different distillation strategies in Table~\ref{tab:ablation_distill}, including no distillation, standard KD, and the proposed TCD.
All variants keep the the proposed framework unchanged and differ only in the distillation strategy.
TCD achieves the best overall performance on both datasets, with an mAP over \textit{all classes} of 93.1 on VOC and 79.9 on TT100K.
Compared with standard KD, TCD improves the mAP over \textit{all classes} by 4.8 points on VOC and 39.8 points on TT100K. These results show that directly applying standard distillation is not sufficient for two-stage DINO. In particular, standard KD decreases the mAP on \textit{base classes} from 92.1 to 88.7 on VOC and reduces the mAP on \textit{incremental classes} from 68.9 to 51.0 on TT100K. This suggests that misaligned supervision can affect both old-class retention and new-class learning. When such misalignment exists, the transferred supervision can be imposed on mismatched regions or object candidates. This not only weakens the preservation of old-class representations, but also conflicts with the learning signals from new-task data, thereby hindering adaptation to new classes. By removing this structural mismatch, TCD provides more reliable supervision in the feature space and improves the balance between old-class retention and new-class adaptation.
This advantage is further supported by the visualization in Fig.~\ref{fig:tsne_compare_all_2x3}.
Without distillation, features of old classes exhibit the most severe displacement after incremental learning, indicating substantial representation drift.
Standard KD alleviates this drift only partially, whereas TCD keeps features of old classes much closer to their original distributions.
For newly learned classes, no distillation produces the most compact clusters, while standard KD and TCD show comparable clustering quality.
These results indicate that TCD achieves a better stability-plasticity trade-off: it substantially improves stability on old classes while maintaining competitive adaptation to newly learned classes.

\subsubsection{Hyperparameter Analysis}
\label{sec:c_ablation}

\begin{figure}[t] 
     \centering
    
     \includegraphics[width=0.85\textwidth]{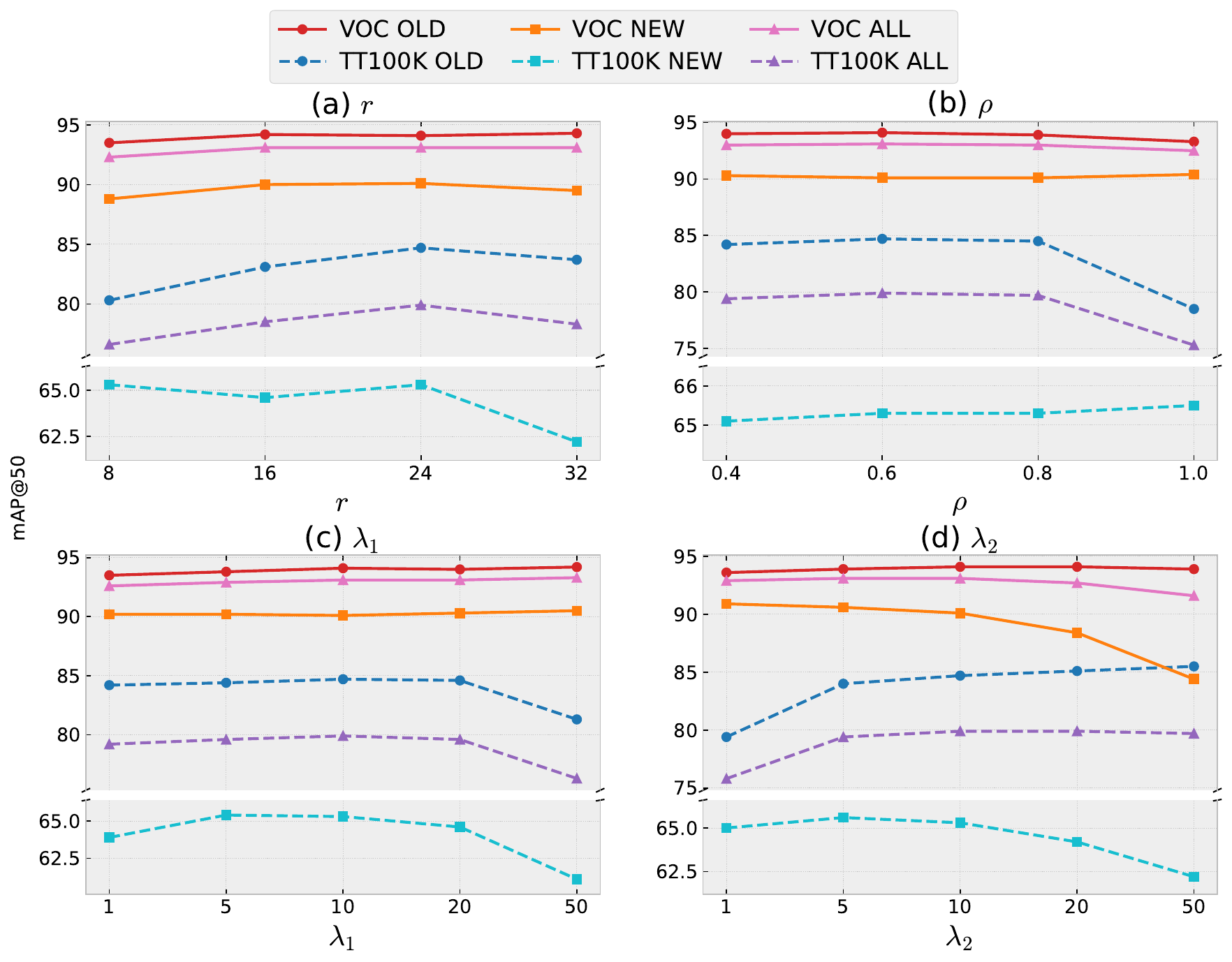}
     \caption{\textbf{Hyperparameter ablations under the 15-5 setting on VOC and TT100K.}
        We study four hyperparameters: the LoRA rank \(r\), the truncation ratio \(\rho\), the replay loss coefficient \(\lambda_1\) and the distillation loss coefficient \(\lambda_2\).
        For each hyperparameter, we report the performance on old, new, and all classes for both VOC and TT100K.}
     \label{fig:hyper_ablation}
\end{figure}

We further analyze the effect of four important hyperparameters in our framework, including the LoRA rank \(r\), the truncation ratio \(\rho\), the replay loss coefficient \(\lambda_1\), and the distillation loss coefficient \(\lambda_2\). 
As shown in Fig.~\ref{fig:hyper_ablation}, the performance on VOC is relatively stable across a moderate range of hyperparameter choices, whereas TT100K is more sensitive and therefore provides clearer guidance for model selection. 

For the LoRA rank \(r\), increasing \(r\) from 8 to 24 consistently improves performance on TT100K, whereas a further increase to 32 provides no additional gain and slightly degrades the overall result. 
This suggests that \(r=24\) offers a suitable balance between adaptation capacity and stability in continual learning.

For the truncation ratio \(\rho\), retaining too many low-confidence features can reduce the reliability of Gaussian estimation, especially on TT100K.
In contrast, \(\rho=0.6\) achieves the best overall performance, supporting our design choice of score-aware feature filtering before Gaussian modeling. 

For the replay loss coefficient \(\lambda_1\), a value that is too small provides insufficient replay supervision, whereas a value that is too large overemphasizes old knowledge.
Interestingly, when \(\lambda_1\) increases from 1 to 5, the new-class performance on TT100K also improves slightly, suggesting that moderate replay can impose a proper constraint on feature adaptation, thereby helping the model learn new classes more effectively.
However, when \(\lambda_1\) becomes too large, the replay constraint starts to hinder plasticity, making \(\lambda_1=10\) the best overall choice. 

A similar phenomenon is observed for the distillation loss coefficient \(\lambda_2\). 
Increasing \(\lambda_2\) from 1 to 5 also brings a slight gain on TT100K new classes, indicating that moderate distillation not only improves old-class stability but can also benefit new-class representation learning. 
However, overly strong distillation suppresses adaptation to new categories.
Among the evaluated values, \(\lambda_2=10\) provides the most favorable balance between stability and plasticity.

\section{Conclusion}

Our work shows that the stability-plasticity trade-off in PTMIOD is not uniform across detector components. 
Pretrained localization heads exhibit strong transferable stability, whereas classification-related representations require plastic adaptation and feature-space retention. 
This asymmetry motivates our localization-classification decoupled framework, where localization-oriented priors are preserved by freezing explicit localization heads, and classification plasticity is enhanced through LoRA-based continual adaptation.
To address old-class forgetting during this adaptation process, we further introduced QGFR and TCD. 
QGFR preserves old-class decision boundaries in the feature space and mitigates forgetting without any exemplars. 
TCD further stabilizes the feature space under continual adaptation, maintaining the reliability of replayed Gaussian distributions and avoiding the misalignment problem of standard distillation in two-stage DETR detectors. 
Extensive results on COCO, VOC, and TT100K demonstrate that these designs jointly achieve a favorable balance between old-class retention and new-class adaptation, leading to SOTA performance under both in-domain and cross-domain IOD settings.

\section{Acknowledgement}
This work was supported by the National Key Research and Development Program of China (Grant No. 2024YFB3309400), the National Natural Science Foundation of China (Grant No. 62277011), the Project of Chongqing MEITC (Grant No. YJX-2025001001009), the Open Research Fund from Guangdong Laboratory of Artificial Intelligence and Digital Economy (SZ) (Grant No. GML-KF-24-18) and CAAI-CANN Open Fund, developed on OpenI Community.

\appendix
\clearpage

\section{More Evidence for Localization Stability Across Domains}
\label{sec:x_localization}
\setcounter{table}{0}
\setcounter{figure}{0}
In Section 4.1, we demonstrated that the localization module in pretrained detector exhibits strong inherent cross-domain stability by evaluating class-agnostic mAR@50 on VOC, COCO and TT100K datasets. 
To further substantiate this finding, we conduct additional experiments on three diverse and more challenging cross-domain datasets: CARPK \cite{hsieh2017drone}, a drone-view dataset for car counting; URPC2019\footnote{www.cnurpc.org}, an underwater dataset for biological object detection; and DeepTrash v2.0 \cite{tata2021robotic}, targeting marine plastic waste detection.
The results, presented in Table~\ref{tab:recall_appendix} as a supplement to Table 1, were obtained using the same experimental setup and evaluation metric.
As shown in the table, the frozen localization module achieves strong performance across all three challenging cross-domain datasets. 
On the CARPK, it achieves 77.3\% mAR@50, a modest 4.7\% drop from its upper bound. 
This stability is even more pronounced on the DeepTrash, where the model reaches a remarkable 96.6\% mAR@50, only a minor 2.6\% drop. 
More notably, even when faced with the severe domain shift of the underwater URPC2019, the model still maintains a high mAR@50 of 82.2\%. 

\begin{figure*}
    
    \centering
    \includegraphics[width=0.98\textwidth]{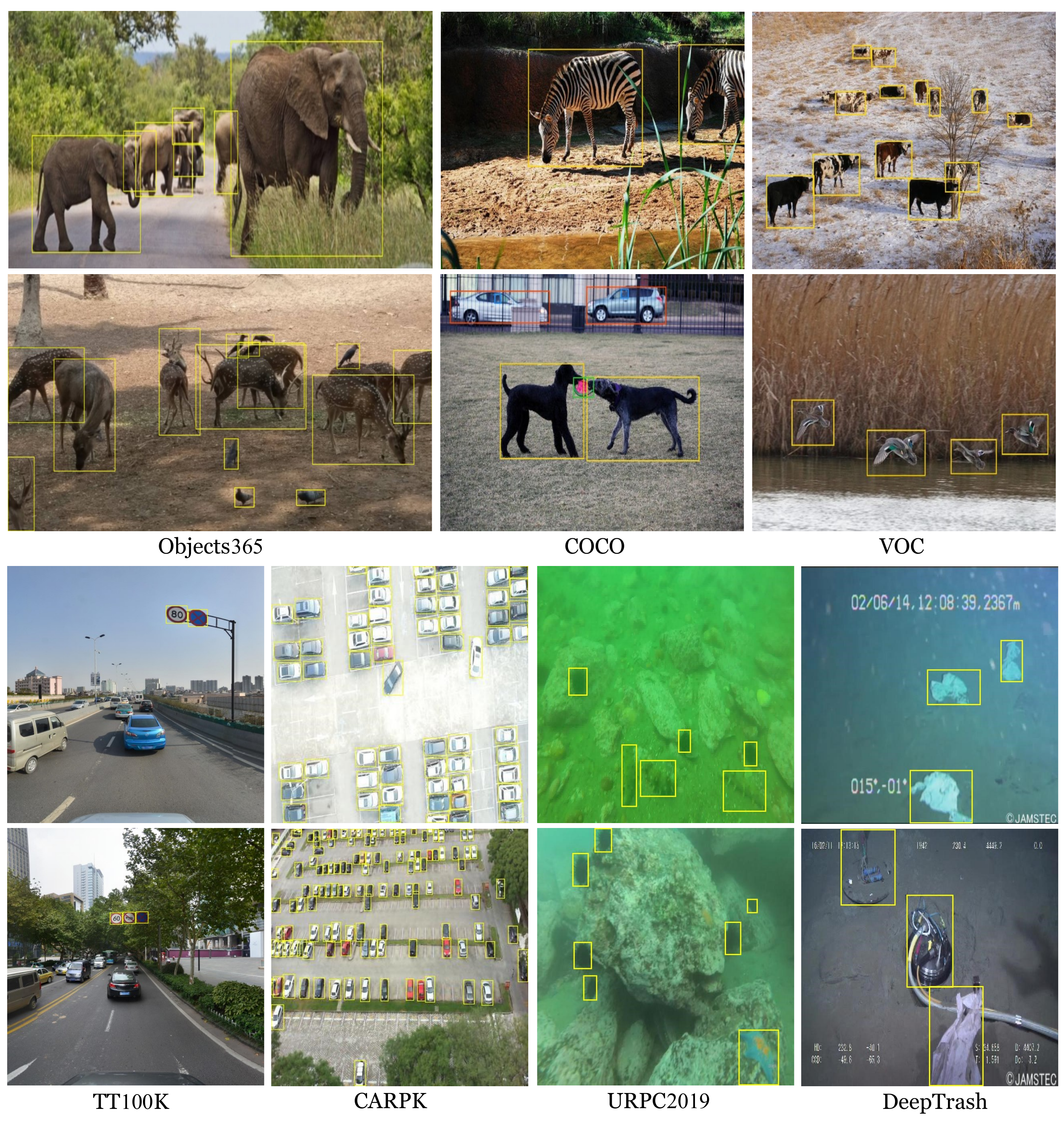}
    \caption{Visualization of domain differences between the pretraining dataset (Objects365) and various datasets mentioned in this work, including both in-domain (COCO, VOC) and cross-domain (TT100K, CARPK, DeepTrash, URPC2019) datasets. Each image is processed using the pretrained detector in a class-agnostic manner to detect object. 
    As we can see in the figure, COCO and VOC exhibit noticeable domain similarity with the pretraining dataset Objects365, whereas TT100K, CARPK, DeepTrash, and URPC2019 differ significantly in terms of visual distribution and content.
    Despite significant visual and distributional discrepancies, the model consistently generates accurate bounding boxes, highlighting the strong cross-domain stability of the localization module in pretrained detector.}

    \label{fig:domain_vis}
    
\end{figure*}

To further provide an intuitive understanding of the domain discrepancies between the pretraining dataset (Objects365 \cite{shao2019objects365}) of PTM and the datasets we mentioned in this work (COCO, VOC, TT100K, CARPK, DeepTrash, and URPC2019), we visualize representative samples in Figure~\ref{fig:domain_vis}. 
For each image, we directly apply the pretrained detector in a class-agnostic setting to detect objects. 
As shown in the figure, COCO and VOC exhibit noticeable domain similarity with the pretraining dataset Objects365, whereas TT100K, CARPK, DeepTrash, and URPC2019 differ significantly in terms of visual distribution and content. 
Despite these substantial domain differences, the pretrained detector consistently produces accurate bounding boxes across all datasets. 

These results collectively demonstrate the strong cross-domain stability of the localization module in pretrained detector, providing support for our proposed modular memory retention strategy.
\begin{table}[t]
\centering
\small 
\caption{Class-agnostic Recall@50 results of the model on datasets from different domains under two training settings. Recall@50 is evaluated at an IoU threshold of 0.5, with all predicted and ground-truth boxes treated as a single class. Lower recall drop indicates better generalization of localization ability.}
\label{tab:recall_appendix}

\setlength{\tabcolsep}{12pt} 

\begin{tabular}{lccc} 
\toprule
\textbf{Dataset} & \textbf{Upper Bound} & \textbf{Frozen Localization} & \textbf{Drop} $\downarrow$ \\ 
\midrule
VOC            & 99.5 & 99.3 & 0.2 \\
TT100K         & 99.7 & 97.3 & 2.4 \\
DeepTrash v2.0 & 99.2 & 96.6 & 2.6 \\ 
CARPK          & 82.0 & 77.3 & 4.7 \\ 
URPC2019       & 97.8 & 82.2 & 15.6 \\
\bottomrule
\end{tabular}
\end{table}

\section{Discussion on Prompt-Based Methods}
\setcounter{table}{0}
\setcounter{figure}{0}
\label{sec:x_prompt}
\begin{figure}[t]
    \centering
    \includegraphics[width=\textwidth]{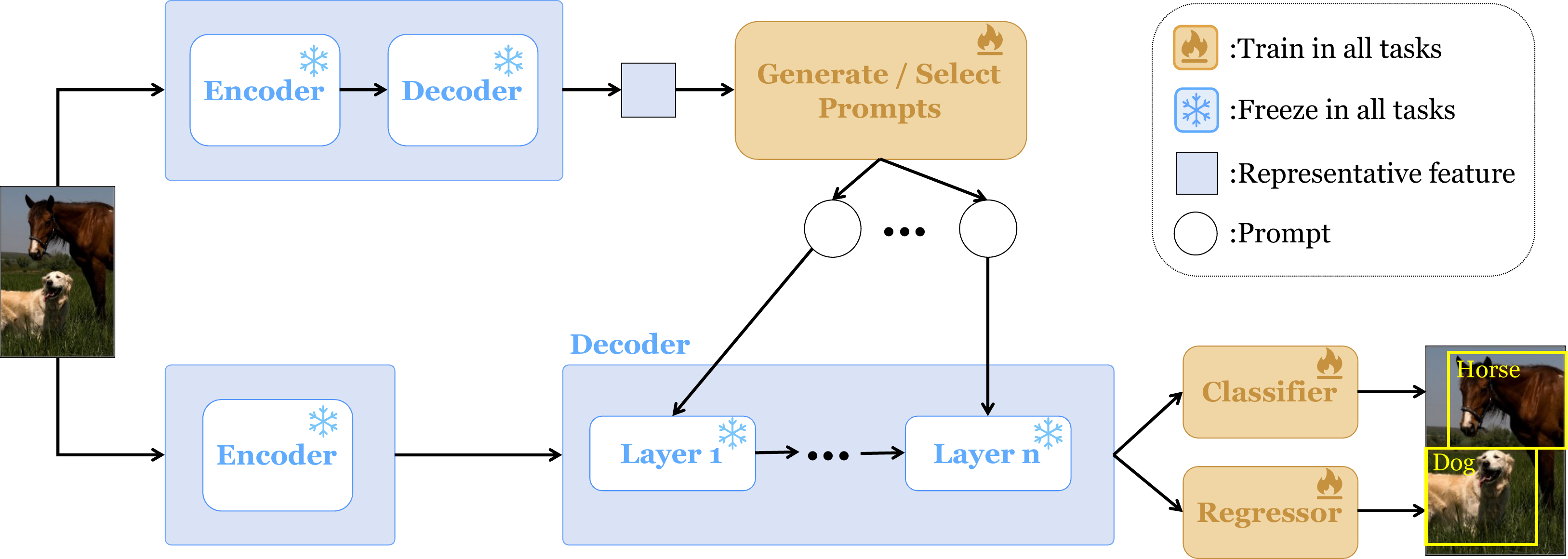}
    \caption{Illustration of a typical prompt-based PTMIOD framework. A frozen pretrained detector first extracts a representative feature, which is then used to generate or select prompts for decoder adaptation.}
    \label{fig:prompt_limit}
\end{figure}

As illustrated in Fig.~\ref{fig:prompt_limit}, existing prompt-based PTMIOD methods typically rely on a frozen pretrained detector to extract a representative feature, which is then used to generate or select prompts.
These prompts are subsequently injected into the decoder to guide downstream adaptation.
This design works well when the downstream data distribution is close to the pretraining domain, because the frozen pretrained detector can still produce stable and discriminative representations.
However, under substantial domain shift, the frozen pretrained detector may fail to produce discriminative features for the new domain.
As a result, the extracted representative feature is no longer sufficiently representative: it becomes noisy, unstable, and exhibits large variance across samples.
As prompt generation or selection is directly conditioned on this feature, the resulting prompts also become unstable and less informative.
This can make optimization on the new task more difficult and slow convergence.
Such inconsistent guidance can also affect continual learning.
When these prompts are not reliably grounded in discriminative representations, they provide limited support for preserving old knowledge, making catastrophic forgetting harder to prevent.
Compared with prompt tuning, LoRA offers a different parameter-efficient strategy: it directly updates the transformation layers of the model, allowing the detector to adjust the representation space for the new domain while keeping the number of trainable parameters small.

\section{Empirical Justification for Gaussian Feature Modeling}
\label{sec:x_Gaussian}
\setcounter{table}{0}
\setcounter{figure}{0}

\begin{figure}[t]
\centering
\includegraphics[width=\textwidth]{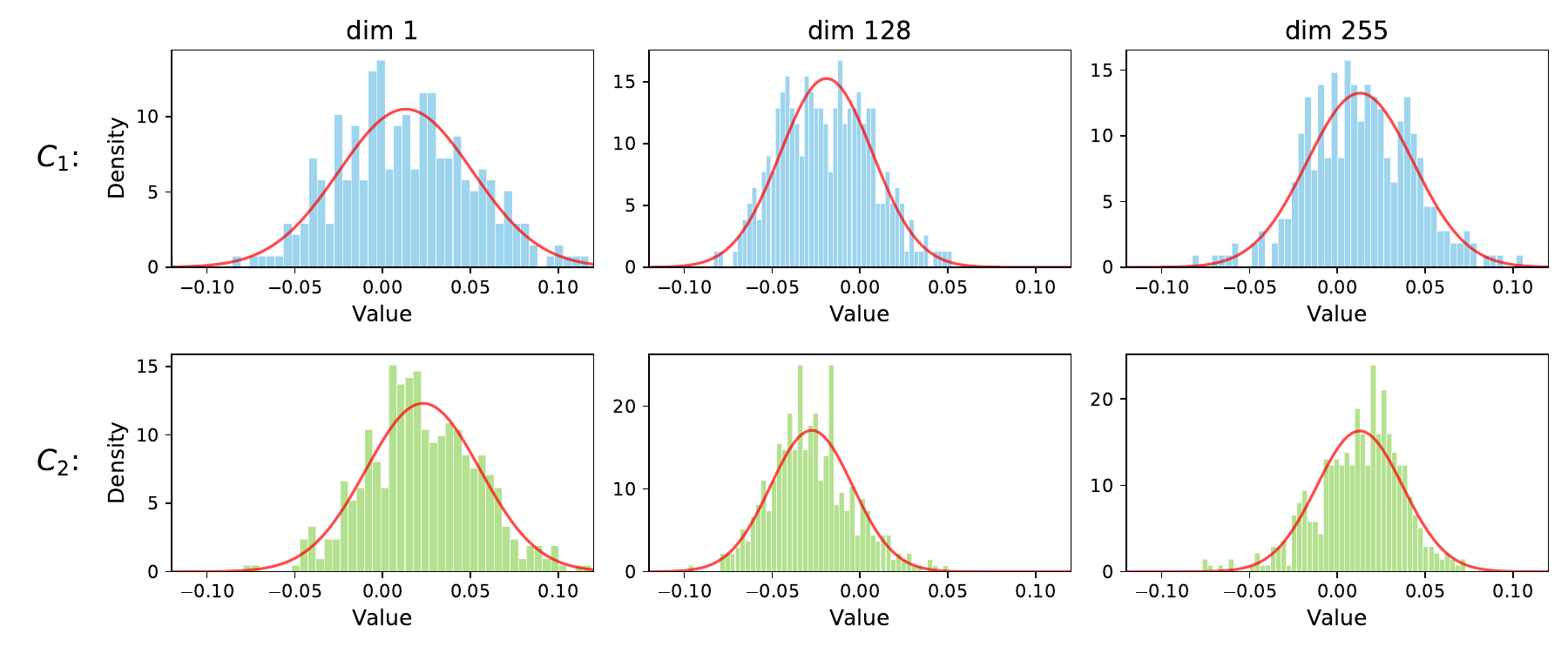}
\caption{\textbf{Evidence for Gaussian feature modeling.}
The empirical histograms of collected class-aligned features are shown together with fitted Gaussian curves for randomly selected categories and representative feature dimensions.}
\label{fig:gaosi}
\end{figure}

Fig.~\ref{fig:gaosi} visualizes the empirical marginal distributions of the collected class-aligned features across several feature dimensions.
After training on task $t$, we collect features for the categories introduced in task $t$ using the feature collection procedure described in Sec. 4.3.1.
From the resulting feature pools, we randomly select two categories and plot the histograms of their feature values along selected dimensions.
For each selected dimension, we fit a Gaussian curve to the corresponding feature values.
As shown in the figure, the fitted Gaussian curves generally match the observed histograms.
This observation suggests that the collected class-aligned features exhibit approximately Gaussian marginal distributions at the individual-dimension level.
It provides empirical support for class-wise Gaussian feature modeling in pseudo-feature replay and is also consistent with prior evidence that pretrained deep feature spaces can be effectively approximated by Gaussian models~\cite{rippel2021modeling}.

\section{Additional Experimental Results}
\label{sec:x_experiment}
\setcounter{table}{0}
\setcounter{figure}{0}

\subsection{Multi-incremental Results on COCO}
Table~\ref{tab:coco_multi_step} reports the multi-increment results on COCO.
Our method consistently outperforms existing approaches under both the 40-10 and 40-20 settings. Compared with P$^2$IOD, it improves the final-step mAP$_{50}$ by 4.3 points under 40-20 and by 6.6 points under 40-10. The larger gain in the latter setting indicates that our method is particularly effective when the detector undergoes more frequent incremental updates.

\subsection{Additional Ablation on Pseudo-Feature Replay}
As shown in Table~\ref{tab:a_ablation}, we further analyze pseudo-feature replay from two complementary perspectives.
Specifically, $\mathcal{L}_{\mathrm{rpy}}^1$ is designed to preserve the ability of the model to extract proposals for previously learned classes, whereas $\mathcal{L}_{\mathrm{rpy}}^2$ aims to preserve the ability of the classifier to recognize old classes.
The full replay loss is defined as $\mathcal{L}_{\mathrm{rpy}}=\mathcal{L}_{\mathrm{rpy}}^1+\mathcal{L}_{\mathrm{rpy}}^2$.

\begin{table}[t]
\centering
\footnotesize
\caption{mAP$_{50}$ on MS COCO under the multi-step settings of 40--20 and 40--10. Results of competing methods are taken from~\cite{an2025parameterized}. Best results are in \textbf{bold}, and second-best results are \underline{underlined}.}
\label{tab:coco_multi_step}
\renewcommand{\arraystretch}{1.12}
\setlength{\tabcolsep}{3pt}
\begin{tabularx}{\linewidth}{>{\raggedright\arraybackslash}p{2.35cm} *{7}{>{\centering\arraybackslash}X}}
\toprule
\multirow{2}{*}{\textbf{Method}} 
& \multirow{2}{*}{\(\mathcal{T}_1\)(1-40)} 
& \multicolumn{2}{c}{\textbf{40+20+20}} 
& \multicolumn{4}{c}{\textbf{40+10+10+10+10}} \\
\cmidrule(lr){3-4} \cmidrule(lr){5-8}
& & \(\mathcal{T}_2\) & \(\mathcal{T}_3\) & \(\mathcal{T}_2\) & \(\mathcal{T}_3\) & \(\mathcal{T}_4\) & \(\mathcal{T}_5\) \\
\midrule
ERD~\cite{feng2022overcoming}          & 63.7 & 54.5 & 48.6 & 53.9 & 46.7 & 39.9 & 31.8 \\
CL-DETR~\cite{liu2023continual}        & 63.7 & 58.3 & 54.1 & 54.4 & 50.2 & 45.6 & 38.2 \\
SDDGR~\cite{kim2024sddgr}              & 68.6 & 62.6 & 59.5 & 62.8 & 60.2 & 59.0 & 54.7 \\
LEA~\cite{song2025learning}            & 75.3 & 62.0 & 57.5 & 66.3 & 61.7 & 59.7 & 56.5 \\
GCD~\cite{wang2025gcd}                 & --   & --   & 60.4 & --   & --   & --   & 55.1 \\
\midrule
MD-DETR~\cite{bhatt2024preventing}     & 79.0 & 69.4 & 60.3 & 68.1 & 61.7 & 53.7 & 49.4 \\
P$^2$IOD~\cite{an2025parameterized}    & 79.6 & 71.3 & \underline{68.8} & 74.1 & 70.9 & 69.3 & \underline{64.8} \\
\rowcolor{gray!10}
Ours                                   & 83.4 & 75.4 & \textbf{73.1} & 77.9 & 73.9 & 73.2 & \textbf{71.4} \\
\bottomrule
\end{tabularx}
\end{table}

\begin{table*}
\caption{Effect of replay loss components on VOC and TT100K. 
All variants use frozen localization heads, LoRA, and TCD, and differ only in the replay loss used during incremental learning.}
\label{tab:a_ablation}
\centering
\centering
\footnotesize
\resizebox{\textwidth}{!}{
\begin{tabular}{l|lll|lll||lll|lll} 
\hline
\multicolumn{1}{l|}{\textbf{Dataset}} & \multicolumn{6}{c||}{\textbf{VOC}} & \multicolumn{6}{c}{\textbf{TT100K}} \\ 
\hline
\textbf{Setting} & \multicolumn{3}{c|}{\textbf{15-5 (2 Tasks)}} & \multicolumn{3}{c||}{\textbf{15-1 (6 Tasks)}} & \multicolumn{3}{c|}{\textbf{15-5 (2 Tasks)}} & \multicolumn{3}{c}{\textbf{15-1 (6 Tasks)}} \\ 
\hline
\textbf{Method} & \textbf{Base} & \textbf{Inc} &  \textbf{All}  & \textbf{Base} & \textbf{Inc} &  \textbf{All} & \textbf{Base} & \textbf{Inc} &  \textbf{All} & \textbf{Base} & \textbf{Inc} &  \textbf{All} \\ 
\hline
LoRA+TCD & 89.3 & 90.1 & 89.5 & 67.6 & 82.1 &  71.3 & 66.2 & 63.6 & 65.6 & 30.7 & 25.8 &  29.5 \\
LoRA+TCD+$\mathcal{L}_{\mathrm{rpy}}^1$ & 92.3 & 90.2 & 91.8 & 67.4 & 82.3 &  71.2 & 66.4 & 63.3 & 65.6 & 52.7 & 28.3 &  46.6 \\
LoRA+TCD+$\mathcal{L}_{\mathrm{rpy}}^2$  & 92.6 & 90.3 &  92.0 & 91.0 & 87.0 & 90.0 & 84.5 & 64.7 & 79.6 &  77.9 & 56.5 & 72.5  \\
LoRA+TCD+$\mathcal{L}_{\mathrm{rpy}}$  & 94.1 & 90.1 &  93.1 & 91.2 & 86.9 & 90.1 & 84.7 & 65.3 & 79.9 & 80.1 & 54.3 & 73.6 \\
\hline
\end{tabular}
}
\end{table*}

First, LoRA+TCD alone is insufficient to preserve old knowledge, especially under long incremental sequences and cross-domain settings.
The severe performance degradation on VOC 15--1 and TT100K indicates that distillation alone cannot adequately maintain either proposal extraction or classification for old classes.

Second, applying only $\mathcal{L}_{\mathrm{rpy}}^1$ mainly benefits proposal preservation for old classes.
This effect is most evident on TT100K 15--1, where performance on both \textit{base classes} and \textit{all classes} improves substantially.
However, the gains under other settings remain limited.
This indicates that preserving old-class proposals alone is not sufficient: even when old-class objects can still be proposed, they may not be correctly recognized if the classifier loses discriminative ability for old classes.

Third, $\mathcal{L}_{\mathrm{rpy}}^2$ plays a more central role in the overall performance.
Compared with $\mathcal{L}_{\mathrm{rpy}}^1$, it yields large gains on both VOC and TT100K, particularly under the more challenging 15--1 settings.
Notably, even without $\mathcal{L}_{\mathrm{rpy}}^1$, $\mathcal{L}_{\mathrm{rpy}}^2$ already achieves performance close to that of $\mathcal{L}_{\mathrm{rpy}}$.
This suggests that the proposal-level classifier is substantially more stable.
Moreover, $\mathcal{L}_{\mathrm{rpy}}^2$ also improves performance on \textit{incremental classes} in several settings, suggesting that jointly training the classifier with replayed old-class pseudo features and current-task features helps maintain clearer decision boundaries between old and new classes.

Finally, $\mathcal{L}_{\mathrm{rpy}}$ consistently achieves the best overall performance across all settings. This shows that $\mathcal{L}_{\mathrm{rpy}}^1$ and $\mathcal{L}_{\mathrm{rpy}}^2$ are complementary: the former helps preserve old-class proposal quality, while the latter preserves old-class classification ability.

\section{Dataset Details}
\setcounter{table}{0}
\setcounter{figure}{0}
\label{sec:x_dataset}
\begin{table}[t]
\centering
\small
\caption{Dataset details and IOD settings.}
\label{tab:dataset_details_appendix}
\resizebox{\linewidth}{!}{%
\begin{tabular}{c|c|c|c||cccc}
\hline
\multirow{2}{*}{\textbf{Dataset}} & \multirow{2}{*}{\textbf{Dataset Details}} & \multirow{2}{*}{\begin{tabular}[c]{@{}c@{}}\textbf{Num of} \\ \textbf{Categories}\end{tabular}} & \multirow{2}{*}{\begin{tabular}[c]{@{}c@{}}\textbf{Obj365} \\ \textbf{Overlap}\end{tabular}} & \multicolumn{4}{c}{\textbf{IOD Settings}} \\
 &  &  &  & 2 Tasks & 3 Tasks & 5 Tasks & 6 Tasks \\ \hline
COCO & \begin{tabular}[c]{@{}c@{}}118,287 train images\\ 5,000 test images\end{tabular} & 80 & 50\% & \multicolumn{1}{c|}{40-40/70-10} & \multicolumn{1}{c|}{40-20} & \multicolumn{1}{c|}{40-10} & - \\ \hline
VOC2007 & \begin{tabular}[c]{@{}c@{}}5,011 train images\\ 4,952 test images\end{tabular} & 20 & 38.9\% & \multicolumn{1}{c|}{15-5} & \multicolumn{1}{c|}{10-5} & \multicolumn{1}{c|}{-} & 15-1 \\ \hline
TT100K & \begin{tabular}[c]{@{}c@{}}5,240 train images\\ 2,586 test images\end{tabular} & 20 & 0.217\% & \multicolumn{1}{c|}{15-5} & \multicolumn{1}{c|}{10-5} & \multicolumn{1}{c|}{-} & 15-1 \\ \hline
\end{tabular}%
}
\end{table}

We select three representative object detection datasets to conduct our experiments, covering both in-domain and cross-domain scenarios: COCO, VOC, and TT100K.  
\paragraph{COCO}  
The COCO dataset contains 118,287 training images and 5,000 validation images, annotated with 80 object categories.
It features complex scenes with multiple objects of varying sizes, occlusions, and diverse contexts, making it a challenging and comprehensive benchmark for general-purpose object detection. 

\paragraph{VOC}  
Pascal VOC 2007 includes 9,963 images with 20 object categories. Although relatively smaller in scale, it remains a standard benchmark due to its high-quality annotations and consistent object categories. We adopt the VOC trainval set for training and the test set for evaluation.

\paragraph{TT100K}  
The Tsinghua-Tencent 100K (TT100K) serves as our primary cross-domain benchmark. It is a large-scale dataset focused on traffic sign detection, containing around 30,000 instances in 100,000 panoramas. 
Due to the long-tailed distribution of class frequencies in TT100K, we select the 20 most frequent traffic sign categories to ensure sufficient training samples and a more stable evaluation. Our experiments are conducted on a subset of 7,826 images covering the following classes: \texttt{i2}, \texttt{i4}, \texttt{i5}, \texttt{il60}, \texttt{io}, \texttt{p10}, \texttt{p11}, \texttt{p26}, \texttt{p5}, \texttt{pl100}, \texttt{pl30}, \texttt{pl40}, \texttt{pl5}, \texttt{pl50}, \texttt{pl60}, \texttt{pl80}, \texttt{pn}, \texttt{pne}, \texttt{po}, and \texttt{w57}.

The first two datasets, COCO and VOC2007, are widely used benchmarks in incremental object detection and exhibit substantial distributional overlap with the pretraining data of our pretrained detector, thus serving as in-domain datasets in our experiments. 
In contrast, TT100K presents a significant domain gap from the pretraining dataset of PTM, and is therefore used as our cross-domain evaluation benchmark. 
To quantify these relationships, we define the overlap as the percentage of instances within Objects365 that belong to the downstream dataset's categories, given that the extensive Objects365 vocabulary largely covers these label sets. This metric represents the proportion of the pretraining domain relevant to the downstream dataset. Based on this, COCO and VOC exhibit high overlaps of approximately 50\% and 38.9\%, respectively, whereas TT100K shows a negligible overlap of about 0.217\%
A summary of these datasets and their relevance to the IOD setting is provided in Table~\ref{tab:dataset_details_appendix}.

\section{Implementation Details}
\label{sec:x_details}
Following P$^2$IOD~\cite{an2025parameterized}, we adopt the same DINO detector architecture with a Swin-L backbone and initialize it with the same weights pretrained on Objects365~\cite{shao2019objects365}.
We train the network using AdamW with a weight decay of \(1\times10^{-4}\). The global batch size is 8. LoRA is inserted into the attention layers, including both self-attention and cross-attention, and its rank \(r\) is fixed to 24 in all experiments. The learning rate of the classification head is set to \(5\times10^{-3}\). For LoRA, we use a learning rate of \(1\times10^{-4}\) on the first task and reduce it to \(1\times10^{-5}\) on subsequent incremental tasks. The truncation ratio \(\rho\) is fixed to 0.6. Both the replay loss coefficient \(\lambda_1\) and the TFD loss coefficient \(\lambda_2\) are set to 10.
On COCO, each incremental task is trained for 1 epoch. On VOC and TT100K, we adopt split-dependent training schedules. For VOC, both tasks in the 15-5 setting are trained for 5 epochs, and all tasks in the 10-5 setting are also trained for 5 epochs. In the 15-1 setting, the five subsequent incremental tasks are trained for 15, 40, 10, 5, and 20 epochs, respectively. For TT100K, the 15-5 setting uses 100 and 15 epochs for the base and incremental tasks, respectively; the 10-5 setting uses 60, 15, and 5 epochs for its three tasks; and the 15-1 setting uses 100, 15, 5, 15, 45, and 50 epochs for its six tasks.
All experiments are conducted on GeForce RTX 3090
GPUs.

\section{Variable Definitions}
\setcounter{table}{0}
\setcounter{figure}{0}
Table~\ref{tab:variable_definitions} summarizes the key notation used in our method.
\begin{table}[t]
\centering
\caption{Summary of key notation used in our method.}
\label{tab:variable_definitions}
\footnotesize
\setlength{\tabcolsep}{4pt}
\renewcommand{\arraystretch}{1.06}
\begin{tabular}{@{}p{0.13\linewidth}p{0.32\linewidth} p{0.13\linewidth}p{0.32\linewidth}@{}}
\toprule
\textbf{Symbol} & \textbf{Definition}
& \textbf{Symbol} & \textbf{Definition} \\
\midrule
$D_t$ & dataset of task $t$
& $C_t$ & classes of task $t$ \\

$C_{1:t-1}$ & old classes
& $\mathcal{M}_t$ & detector after task $t$ \\

$(y_i,\mathbf{b}_i)$ & ground-truth label and box
& $(\hat{\mathbf{b}}_i,\hat{c}_i)$ & predicted box and label \\

$\mathbf{Z}^{1},\mathbf{Z}^{2}$ & encoder and decoder features
& $\mathbf{z}_i^{*}$ & feature token at stage $*$ \\

$* \in \{1,2\}$ & stage index
& $\mathcal{I}$ & Top-$k$ indices \\

$\mathcal{F}^{1},\mathcal{F}^{2}$ & classification heads
& $\mathcal{G}^{1},\mathcal{G}^{2}$ & box regressors \\

$Q^{\mathrm{pos}},Q^{\mathrm{con}}$ & decoder queries
& $W,\Delta W$ & weight and LoRA update \\

$A,B,r$ & LoRA matrices and rank
& $\hat{\sigma}^{*}$ & matching assignment \\

$\mathcal{P}_{t,c}^{*}$ & candidate feature pool 
& $\widetilde{\mathcal{P}}_{t,c}^{*}$ & filtered feature pool \\

$q_i^{*}$ & feature quality score
& $\rho$ & retention ratio \\

$\boldsymbol{\mu}_{t,c}^{*},\boldsymbol{\Sigma}_{t,c}^{*}$ & Gaussian statistics
& $\mathcal{N}(\boldsymbol{\mu}_{t,c}^{*},\boldsymbol{\Sigma}_{t,c}^{*})$ & replay distribution \\

$\tilde{\mathbf{z}}_{i,c}^{*}$ & pseudo-feature
& $\tilde{\mathcal{Z}}^{*}$ & pseudo-feature set \\

$N_t,S_t$ & instances and iterations
& $\mathcal{I}_{\mathrm{tc}}$ & teacher Top-$k$ indices \\

$\mathbf{Z}_{\mathrm{stu}}^{1},\mathbf{Z}_{\mathrm{tc}}^{1}$ & student and teacher features
& $\lambda_1,\lambda_2$ & loss weights \\

$\mathcal{L}_{\mathrm{DINO}}$ & DINO loss
& $\mathcal{L}_{\mathrm{rpy}}$ & replay loss \\

$\mathcal{L}_{\mathrm{tcd}}$ & TCD loss
& $\mathcal{L}_{\mathrm{total}}$ & total loss \\
\bottomrule
\end{tabular}
\end{table}

\bibliographystyle{elsarticle-num}
\bibliography{ref}
\end{document}